\documentclass[lettersize,journal]{IEEEtran}

\usepackage{amsmath,amsfonts,amssymb,amsthm}
\usepackage{array}
\usepackage[caption=false,font=normalsize,labelfont=sf,textfont=sf]{subfig}
\usepackage{textcomp}
\usepackage{stfloats}
\usepackage{url}
\usepackage{verbatim}
\usepackage{graphicx}
\usepackage{cite}
\usepackage{hyperref}
\usepackage[ruled,vlined,linesnumbered]{algorithm2e}
\usepackage{booktabs}
\usepackage[table]{xcolor}
\usepackage{tikz}
\usepackage{multirow}
\usepackage{makecell}
\usepackage{contour}
\usepackage{float}
\usepackage{dashrule}
\usepackage{tabularx}
\usepackage{etoolbox}
\usepackage{pifont}
\usepackage{environ}

\theoremstyle{definition}

\def\BibTeX{{\rm B\kern-.05em{\sc i\kern-.025em b}\kern-.08em
    T\kern-.1667em\lower.7ex\hbox{E}\kern-.125emX}}

\NewEnviron{subfigure}[2][]{%
  \def\subfigcaption{}%
  \long\def\caption##1{\gdef\subfigcaption{##1}}%
  \subfloat[\subfigcaption]{%
    \begin{minipage}[#1]{#2}
      \centering
      \BODY
    \end{minipage}
  }%
}

\newif\ifanonymous
\anonymoustrue

\definecolor{mycyan}{RGB}{253,242,205}
\definecolor{myyellow}{RGB}{255,217,102}
\definecolor{myPink}{RGB}{199,121,152}
\definecolor{myMint}{RGB}{217,234,211}
\definecolor{mysoftSage}{RGB}{219,234,211}
\definecolor{softCoral}{RGB}{227,167,154}
\definecolor{lightGrayish}{RGB}{236,234,232}

\ifanonymous
  
  \newcommand{\xuehui}[1]{#1}
  \newcommand{\yihui}[1]{#1}
  \newcommand{\rushi}[1]{}
\else
  
  \newcommand{\xuehui}[1]{{#1}}
  \newcommand{\yihui}[1]{{#1}}
  \newcommand{\rushi}[1]{\textcolor{magenta}{[Rushi: #1]}}
\fi

\contourlength{0.08em}

\newcommand{\unsafeText}[1]{\contour{black}{\textcolor{myPink}{\textbf{#1}}}}
\newcommand{\safeText}[1]{\contour{black}{\textcolor{softCoral}{\textbf{#1}}}}
\newcommand{\clearText}[1]{\contour{black}{\textcolor{lightGrayish}{\textbf{#1}}}}

\newcommand{\navYes}{\contour{black}{\textcolor{white}{\ding{51}}}}
\newcommand{\navNo}{\contour{black}{\textcolor{white}{\ding{55}}}}

\DeclareRobustCommand{\colorcube}[1]{%
  \begingroup
  \setlength{\fboxsep}{0pt}%
  \colorbox{#1}{\phantom{\rule{1.8ex}{1.8ex}}}%
  \endgroup
}

\renewcommand{\arraystretch}{1.15}
\newcolumntype{P}[1]{>{\raggedright\arraybackslash}p{#1}}
\newcolumntype{C}[1]{>{\centering\arraybackslash}p{#1}}

\title{Parallel OctoMapping: A Scalable Framework for Enhanced Path Planning in Autonomous Navigation}

\ifanonymous

\author{Yihui Mao, Tian Tan, Xuehui Shen, Warren E. Dixon, and Rushikesh Kamalapurkar%
\thanks{This research is supported, in part, by the Air Force Research Laboratory under Grant No.~FA8651-24-1-0019, the Air Force Office of Scientific Research under Grant No.~FA9550-19-1-0169, and the U.S. Army Research Laboratory under Grant No.~W911NF-25-2-0045. Any opinions, findings, and conclusions or recommendations expressed in this material are those of the author(s) and do not necessarily reflect the views of the sponsoring agencies.}%
\thanks{Yihui Mao, Xuehui Shen, Warren E. Dixon, and Rushikesh Kamalapurkar are with the Department of Mechanical and Aerospace Engineering, University of Florida, Gainesville, FL 32611 USA (e-mail: \{yihui.mao, xuehuishen, wdixon, rkamalapurkar\}@ufl.edu).}%
\thanks{Tian Tan is with the Department of Electrical and Systems Engineering, University of Pennsylvania, Philadelphia, PA 19104 USA (e-mail: tiantan@alumni.upenn.edu).}%
}

\fi
\begin{document}

\maketitle

\begin{abstract}
Mapping is essential in robotics and autonomous systems because it provides the spatial foundation for path planning. Efficient mapping enables planning algorithms to generate reliable paths while ensuring safety and adapting in real time to complex environments. Fixed-resolution mapping methods often produce overly conservative obstacle representations, which can lead to suboptimal paths or planning failures in cluttered scenes. To address this issue, we introduce Parallel OctoMapping (POMP), an efficient OctoMap-based mapping technique that preserves more navigable free space and supports multi-threaded computation. To the best of our knowledge, POMP is the first method that refines the representation of free space at a fixed occupancy-grid resolution without changing the underlying grid structure, while preserving compatibility with existing search-based planners. It can therefore be integrated into existing planning pipelines, yielding higher pathfinding success rates and shorter path lengths, especially in cluttered environments, while substantially improving computational efficiency. An interactive web-based demonstration illustrating the mapping and planning behavior of POMP is available on the project webpage.\footnote{\url{https://maoyihui53.github.io/pomp-demo/}}
\end{abstract}


\section{INTRODUCTION}

\yihui{Understanding the environment is fundamental to autonomy in mobile robotics.} A well-designed mapping method that enforces spatial consistency across observations, produces an accurate and memory-efficient representation of the environment, and supports fast, informative spatial queries \yihui{is thus a critical component of modern autonomy systems.} \yihui{Therefore, good map representations improve the reliability of downstream planning and collision-avoidance modules, enabling safer, higher-quality trajectories and ultimately more robust autonomous navigation~\cite{hornung2013octomap, KinectFusion, 6599048, SCC.Qureshi.Ogri.ea2024, thrun2002probabilistic}.}

In online navigation, \yihui{mapping }accuracy alone is insufficient. The mapping module must also meet \yihui{time and computation} constraints, supporting fast construction and frequent updates without sacrificing fidelity (runtime efficiency), while keeping memory usage bounded through sparse representations (memory efficiency). \yihui{During execution, the map should provide online querying capability, allowing the planner to obtain the spatial information it needs, so that mapping and planning can work together within a unified navigation pipeline (continuous usability).} OctoMap is a widely adopted representation for 3D occupancy mapping because its sparse octree structure enables memory-efficient storage, multi-resolution queries, and incremental updates~\cite{hornung2013octomap}. However, in large environments with dense point clouds and high update rates, a standard OctoMap pipeline can become compute-bound and may struggle to sustain real-time throughput.

Search-based methods are widely used for planning because they integrate naturally with grid and voxel maps and can provide completeness and optimality guarantees under standard assumptions~\cite{dijkstra2022note, hart1968formal, stentz1994optimal, harabor2011online}. The performance of search-based planning, however, is strongly influenced by map resolution and \yihui{the choice of representation.} \yihui{With fixed resolution cells, typical occupancy grids may mark an entire cell as occupied when a relatively small portion of an obstacle falls within it.} \yihui{The conservative labeling rule can unnecessarily mark traversable space as non-traversable, resulting in broken connectivity in narrow passages, increased search effort, degraded path quality, and in some cases planning failures in cluttered scenes~\cite{latombe1991robot}.} A straightforward remedy is \yihui{to increase the grid resolution, but this substantially increases mapping computation and memory consumption, enlarges the search graph, and ultimately lengthens planning time.} \yihui{Motivated by these limitations, we seek map representations that better exploit within-cell free space to preserve narrow passages without globally refining the resolution, while remaining efficient for online planning.}

To address the limitations of fixed-resolution mapping for path planning, we propose an efficient mapping technique based on OctoMap \yihui{that} accelerates map construction through multi-threaded parallel computation and maximizes the utilization of cell space at fixed resolution. While conventional methods conservatively label an entire cell as occupied whenever it contains even a small portion of an obstacle, Parallel OctoMapping (POMP) performs a finer-grained analysis of the internal clustered spatial distribution of the point cloud to safely reclaim significant navigable space that would otherwise be inaccessible under traditional methods. In summary, our contributions are as follows:


\begin{itemize}
    \item A novel OctoMap-based mapping technique (POMP) \yihui{is developed} that improves grid/voxel space utilization in 2D and 3D \yihui{by subdividing each fixed-resolution grid/cell into distinct sub-regions}. \yihui{By introducing  \texttt{clear},  \texttt{safe}, and  \texttt{unsafe} states based on the clustered spatial distribution of the point cloud,} this method unlocks significant navigable space overlooked by conservative Occupancy Grid Map (OGM) methods.
    \item POMP performs parallel, multi-threaded computation, significantly reducing map construction time in large environments with dense point clouds and frequent updates.
    \item POMP improves pathfinding success rate and path quality of search-based path planning while reducing mapping time compared with conventional fixed-resolution occupancy grid methods.
\end{itemize}

\section{related work}

\subsection{Mapping Representations and Frameworks}
Point clouds, commonly captured in robotic tasks like mapping and planning, often contain excessive points and sensor noise, making them less suitable for efficiently representing large-scale environments. One simple alternative is the grid/voxel map, which partitions space into uniform squares/cubes to represent the scene; however, many of these remain unexplored \yihui{in each sensor measurement}, leading to \yihui{a} substantial memory overhead. 

Tree-based representations have been studied to overcome these issues. The \yihui{OctoMap} ~\cite{hornung2013octomap} is a well-established representation that \yihui{uses an Octree}, which divides  \yihui{ 3D} space into eight subspaces \yihui{that have the same volume.}  \yihui{OctoMaps compactly and probabilistically represent an environment with occupancy states, including unknown, occupied, or free.}

In recent years, a variety of mapping frameworks \cite{liu2023dataframe, duberg2020ufomap, sun2018recurrentoctomap, Chen.Li.Wan.ea2025, SCC.Kwon.Kim.ea2019, SCC.Min.Han.ea2023.OctoMap-RT, hornung2013octomap,  cai2024occupancy} have been proposed to improve map representation and computational efficiency. In addition, some of these frameworks \cite{oleynikova2017voxblox,Han.Gao.Zhou.ea2019,Pan.Kompis.Bartolomei.ea2022} facilitate navigation by providing map representations that are more suitable for path planning. Truncated Signed Distance Fields (TSDFs) ~\cite{curless1996volumetric,KinectFusion}, originally used in computer graphics, represent geometry implicitly by \yihui{storing truncated signed distance to the observed surface}. In practice, TSDF values are updated using distances projected along the sensor ray, and they are maintained only within a narrow truncation band around the surface. Voxblox ~\cite{oleynikova2017voxblox} builds upon \yihui{TSDF} by incrementally constructing Euclidean Signed Distance Fields (ESDFs) from the TSDF map, allowing efficient queries of the Euclidean distance from each voxel to the nearest obstacle to support path planning. FIESTA ~\cite{Han.Gao.Zhou.ea2019} is another well-known fast incremental ESDF mapping framework for online motion planning of aerial robots. It computes the ESDF directly from an occupancy grid map and builds a growing global map, \yihui{and has been reported to achieve higher accuracy and computational performance compared to Voxblox in its experiments.}

\subsection{Planning on Structured Maps}

Discrete volumetric occupancy maps are typically implemented using one of three commonly used structures: uniform voxels~\cite{7839930,zhou2020ego,liu.mao.belta2025acc}, Octrees (e.g., OctoMap~\cite{hornung2013octomap}), and hashed voxel/block structures ~\cite{oleynikova2017voxblox,niessner2013real}.

Occupancy grid mapping, where uniform voxel cells discretize the space, is easy to implement and aligns well with data from LiDAR and RGB-D cameras; however, \yihui{dense implementations allocate memory uniformly across the bounded map volume, which leads to rapidly increasing memory usage as the resolution becomes finer.}
Octree–based representations such as OctoMap provide hierarchical, memory–efficient storage
and naturally support multi–resolution updates, but their pointer–based hierarchy incurs
$O(\log n)$ access time and makes neighborhood queries across different depths cumbersome. 
Hashed voxel/block structures can achieve \yihui{sparse access in expected near} $O(1)$ access time and scale well to large
environments, but suffer from potential hash collisions and require careful memory
management.

To combine the strengths of these approaches while avoiding their weaknesses, we use the octree solely as a parallel point cloud reader and storage backend, streaming its leaf states to a fixed-resolution occupancy grid in real time. Atomic updates propagate each leaf change to the corresponding grid cell without race conditions, so the planner operates on a grid array with \yihui{ per-voxel access  time of }$O(1)$ \yihui{ and }constant-stride neighbor checks, without traversing the tree during planning, while the map remains sparsely represented in the tree. This decoupled design preserves the octree’s efficient incremental construction and provides the planner with a contiguous grid array for fixed-stride access, enabling fast search-based planning.

\section{Preliminaries}
\subsection{OctoMap and Occupancy Grid Mapping Overview } \label{sec:conf}
OctoMap~\cite{hornung2013octomap} is a 3D occupancy mapping framework that uses an Octree-based data structure to efficiently store, update, and query volumetric information. An Octree is a hierarchical structure that recursively subdivides 3D space into cubic volumes, or voxels, starting from a root node and continuing until a predefined resolution is reached.

The occupancy grid map for pathfinding is generated from the Octree built in the mapping process. The OctoMap and \yihui{the} occupancy grid map are configured as follows. 

\medskip
\noindent\textbf{\yihui{Octree Configuration.}}
The first step is to align the OctoMap with the workspace of interest. We set the root center $\mathbf{c}_{\text{root}}$ of the Octree to coincide with the geometric center 
$\mathbf{c}_{\text{map}}$ of the environment bounds along each coordinate axis. 
Let $L_x, L_y, L_z$ denote the side lengths of the workspace along the $x$, $y$, and $z$ axes. 
We then define $L_{\max}=\max\{L_x, L_y, L_z\}$ as the maximum extent of the environment. 
The depth $n$ of the Octree is determined by the smallest  resolution $r$ (i.e., the edge length of a leaf cube). 
Since each additional tree level subdivides a cube into 8 smaller ones, the minimum depth required to achieve a leaf size $r$ is computed as
\[
n=\left\lceil \log_2\!\left(\frac{L_{\max}}{r}\right)\right\rceil.
\]
The depth guarantees that the root node has a side length 
$S_0 = 2^{\,n} r$ which is large enough to cover the maximum \yihui{workspace} dimension $L_{\max}$ in all directions. 
Intuitively, this step defines the “outer box” of the Octree and ensures that the \yihui{leaf nodes} match the desired map resolution.

\medskip
\noindent\textbf{\yihui{ Grid Configuration.}} 
\yihui{$N_x$, $N_y$, and $N_z$ denote the numbers of leaf nodes along the $x$, $y$, and $z$ axes after discretizing the half of the workspace with leaf size $r$:
\[
N_x=\Big\lceil \tfrac{L_x}{2r} \Big\rceil,\quad
N_y=\Big\lceil \tfrac{L_y}{2r} \Big\rceil,\quad
N_z=\Big\lceil \tfrac{L_z}{2r} \Big\rceil,
\]
where $L_x$, $L_y$, and $L_z$ are the workspace extents along the $x$, $y$, and $z$ axes, respectively.}

\yihui{
The origin of the occupancy grid, $\mathbf{o}_{\text{grid}}$, is placed at the vertex with the smallest coordinates in all three dimensions: 
}
\yihui{
\[
\mathbf{o}_{\text{grid}}
=
\begin{bmatrix}
\mathbf{c}_{\text{map}_x}-N_x r -\tfrac{r}{2}\\
\mathbf{c}_{\text{map}_y}-N_y r -\tfrac{r}{2}\\
\mathbf{c}_{\text{map}_z}-N_z r -\tfrac{r}{2}
\end{bmatrix}.
\]
Finally, the grid dimensions are defined as:
\[
\dim_x = 2N_x+1,\quad \dim_y = 2N_y+1,\quad \dim_z = 2N_z+1,
\]}
\yihui{The grid configuration guarantees that the occupancy grid is symmetrically aligned with the octree root, spans the full environment, and extends by half a cell $(r/2)$ beyond the outermost octree leaf nodes.}




\subsection{\yihui{Parallelization of octree construction}}

\yihui{A} dense point cloud input can make OctoMap construction a computational bottleneck, given that the commonly used implementation \yihui{is often used in a single-threaded or effectively sequential manner in practice}. \yihui{Sequential processing} limits throughput and responsiveness in real time mapping tasks, especially when handling large-scale or high-frequency sensor data. In addition, the hierarchical nature of the octree makes fine-grained parallelization difficult, because both updates and queries require root-to-leaf traversal, and concurrent operations may contend for shared nodes, necessitating synchronization to avoid race conditions.

Existing efforts to accelerate OctoMap can be broadly divided into hardware-assisted and software-based approaches. Hardware solutions, such as OMU \cite{SCC.Jia.Yang.ea2022}, improve performance but reduce portability across the heterogeneous computing platforms used in robotic systems. Other hardware-assisted methods, including NanoMap \cite{SCC.Florence.Carter.ea2018}, OctoMap-RT \cite{SCC.Min.Han.ea2023.OctoMap-RT}, OctoMap build based on Super Rays \cite{SCC.Kwon.Kim.ea2019} and GPU-accelerated OctoMap \cite{SCC.Min.Han.ea2021accel}, mainly accelerate ray tracing, yet octree updates can still remain the dominant cost. Software approaches are more limited. VoxelCache \cite{SCC.Durvasula.Kiguru.ea2023} reduces voxel-access latency through on-chip caching of recently used voxel-block pointers, but it does not parallelize OctoMap tree construction. SkiMap \cite{SCC.DeGregorio.DiStefano2017} improves efficiency by replacing the octree with a Tree of SkipLists and is thus better regarded as an alternative mapping structure than a direct optimization of OctoMap. Among software methods that preserve the OctoMap octree, OctoCache \cite{Chen.Li.Wan.ea2025} is the closest, although its gains mainly arise from caching and workflow-level concurrency rather than true parallel updates within a single Octree.

We parallelize OctoMap construction with Intel oneAPI Threading Building Blocks \yihui{(}oneTBB\yihui{)}, a task-based C++ library whose work-stealing scheduler balances the irregular, fine-grained tasks in voxel updates\cite{reinders2007intel}. Compared with GPU approaches~\cite{SCC.Min.Han.ea2023.OctoMap-RT,SCC.Min.Han.ea2021accel}, CPU parallelism \yihui{can} avoid host–device transfer costs when data reside on the CPU and often better matches the branching and sparse access patterns of Octree updates. Among CPU options, both OpenMP and oneTBB support dynamic scheduling and task parallelism. We choose oneTBB for its composable task-graph API and work-stealing runtime, which can better accommodate many small, irregular tasks than a simple thread-pool design.

To ensure correctness during concurrent updates, we employ the compare-and-swap (CAS) instruction, an atomic primitive supported by most modern multiprocessor architectures. CAS atomically compares the value at a memory location with an expected value and, only if they match, updates it to a new value. During parallel octree construction, when multiple threads concurrently attempt to create the same child node, each thread allocates a candidate node and uses CAS to atomically update the corresponding child pointer from null to that node. If the CAS fails, the thread discards the candidate node and proceeds with the child that has already been created by another thread. Each successful CAS thus ensures that only one thread initializes a given child pointer, preventing duplicate node creation under contention. Compared with the mutex-based locking used in OctoMap, this CAS-based design enables fine-grained child-node creation, reducing lock contention, thread blocking, and context-switch overhead, thereby improving the parallel throughput and scalability of hierarchical octree updates.

\begin{figure}[h!]
    \centering
    \includegraphics[width=1.0\linewidth]{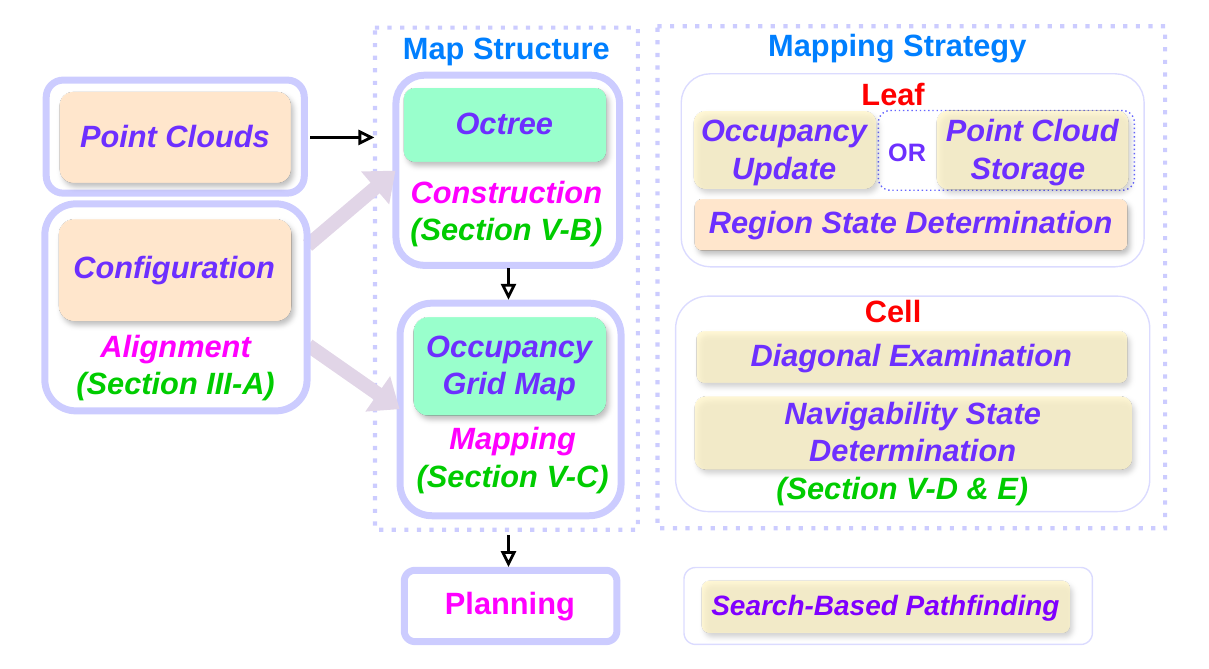}
    \caption{System overview of our proposed mapping framework }
    \label{fig:diagram}
\end{figure}

\section{Overview}

Fig.~\ref{fig:diagram} provides an overview of the proposed POMP framework. The framework consists of two tightly coupled components: (i) an OctoMap backend for real-time map construction, and (ii) a planner-facing fixed-resolution occupancy grid map (OGM) to support efficient search-based planning. Specifically, the OctoMap is continuously updated online, and the leaf-level region states are projected onto the OGM to determine the navigability of each corresponding grid cell for planning.

\begin{figure}[h!]
    \centering
    \includegraphics[width=1.0\linewidth]{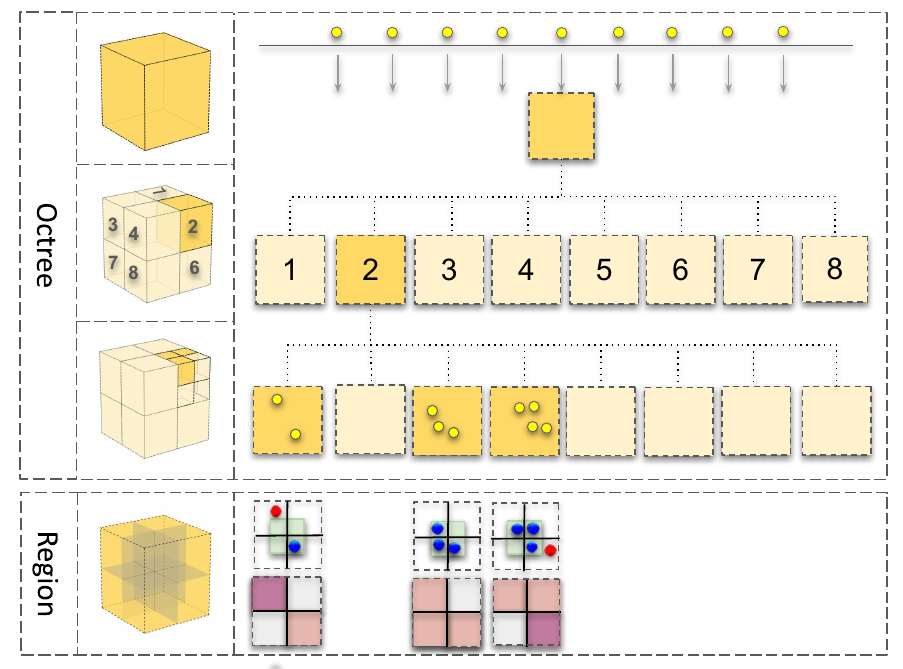}
\caption{Parallel insertion of point clouds into the Octree with concurrent node construction. Each leaf node is subdivided into 4 regions in 2D or 8 regions in 3D. \yihui{In the figure, `` \colorcube{myyellow} '' represents occupied, `` \colorcube{mycyan} '' represents unoccupied, and `` \colorcube{myMint} '' denotes the threshold bounding box. The bounding box is centered on a leaf node and used to determine the region state; by default, the edge length of the bounding box is set to half the leaf size.}  The regions are classified as \yihui{\texttt{clear} `` \colorcube{lightGrayish} ''} (no points),  \yihui{\texttt{safe} `` \colorcube{softCoral} '', or \texttt{unsafe} `` \colorcube{myPink} '' (see Fig.\ref{fig:trd}).}
}

    \label{fig:octomap_build}
\end{figure}

Because the OctoMap and the OGM are tightly coupled, their spatial relationship is explicitly defined (see Section~III-A). In particular, the OctoMap leaf-node size is set equal to the OGM cell \yihui{size}. The number of OGM cells is \yihui{selected to be} larger by one along each axis than the number of OctoMap leaf nodes, and it is shifted by a half-cell offset so that the two discretizations are staggered by half of the cell size. This arrangement yields a consistent, overlap-based correspondence between OctoMap leaf nodes and OGM cells, enabling reliable projection of leaf-level region states onto the grid for planning. \yihui{Due to half-cell offset, the leaf-node boundaries at the chosen depth often pass through the centers of OGM cells, effectively slicing the original grid and introducing additional split lines; because these boundaries are uniformly spaced, any narrow passage of sufficient width (larger than the chosen OGM cell resolution) must be intersected by at least one boundary (pigeonhole principle) and therefore cannot be “skipped” by discretization. We then apply occupancy thresholding with a safety margin to preserve the required clearance even when an obstacle occupies only a small fraction of an OGM cell (See the Region block in Fig.~\ref{fig:octomap_build} and the left panel of Fig.~\ref{fig:trd}).}
POMP starts by acquiring a stream of point clouds, either directly from range sensors (e.g., LiDAR) or from an existing point cloud map. An OctoMap is then constructed with a predefined leaf resolution, where incoming measurements are integrated through leaf-node point storage and/or occupancy updates \yihui{(see Section~\ref{sec:octree construction})}. Based on the resulting leaf-level statistics, \yihui{we assign each leaf-node region a navigability label (\texttt{unsafe} / \texttt{safe} / \texttt{clear}) based on occupancy thresholding, and use this label to evaluate the traversability of the corresponding OGM cell, as discussed in Sections~\ref{sec:mapping} and~\ref{sec:DiagonalExamination}.} 
\begin{table}[t!]
\centering
\caption{Terminology and notation.}
\label{tab:terminology}
\scriptsize
\renewcommand{\arraystretch}{1.2}
\setlength{\tabcolsep}{2pt}

\begin{tabular}{|
  >{\centering\arraybackslash}m{0.15\columnwidth}|
  >{\raggedright\arraybackslash}m{0.14\columnwidth}|
  >{\raggedright\arraybackslash}m{0.55\columnwidth}|
  >{\centering\arraybackslash}m{0.07\columnwidth}|
}
\hline
\textbf{Category} & \textbf{Term} & \textbf{Definition} & \textbf{Sym.} \\
\hline
\multirow{6}{*}[-1.5\baselineskip]{OctoMap}
& node & A hierarchical spatial unit in the Octree corresponding to a cubic region of space. & \tikz{\draw[dash pattern=on 3.5pt off 1.2pt, line width=0.35pt] (0,0) -- (0.85\linewidth,0);} \\
\cline{2-4}
& leaf size & The edge length of a leaf node. & $r$ \\
\cline{2-4}
& threshold & A user-defined range to classify points as unsafe, safe or clear for navigation. & thr \\
\cline{2-4}
& occupied state & A state in which the OctoMap node contains point cloud. & \colorcube{myyellow} \\
\cline{2-4}
& unoccupied state & A state where no point cloud is marked in an OctoMap node. & \colorcube{mycyan} \\
\hline
\multirow{4}{*}[-1.6\baselineskip]{Region}
& region & A leaf node equally separated into 4 regions in 2D and 8 regions in 3D. & \rule{0.90\linewidth}{0.4pt} \\
\cline{2-4}
& unsafe state & A region state in a leaf node with points outside the range set by the threshold. & \colorcube{myPink} \\
\cline{2-4}
& safe state & A region whose points remain within the threshold bound and are treated as axis-aligned traversable. & \colorcube{softCoral} \\
\cline{2-4}
& clear state & A region state in a leaf node without a point cloud. & \colorcube{lightGrayish} \\
\hline
\multirow{4}{*}[0.1\baselineskip]{\parbox{0.15\columnwidth}{\centering Occupancy\\Grid Map}}
& grid & A discrete unit of the 2D occupancy grid map. & \tikz[baseline=-0.6ex]{\draw[dash pattern=on 2.2pt off 1.0pt on 0.9pt off 1.0pt, line width=0.4pt] (0,0) -- (0.90\linewidth,0);} \\
\cline{2-4}
& voxel & A discrete unit of the 3D occupancy grid map. & \tikz[baseline=-0.6ex]{\draw[dash pattern=on 2.2pt off 1.0pt on 0.9pt off 1.0pt, line width=0.4pt] (0,0) -- (0.90\linewidth,0);} \\
\cline{2-4}
& resolution & The grid/voxel edge length, same as the leaf size in our method. & res \\
\hline
\end{tabular}
\end{table}

\yihui{The developed method} fundamentally overcomes a key limitation of existing mapping techniques. The conventional methods \cite{30720,1087316} employ overconservative occupancy labeling, marking a grid cell as fully occupied regardless of whether it is densely filled or \yihui{contains} only a few sparse points. \yihui{Our approach, described in Algorithm~\ref{alg:parallel_octoMap_build} and Algorithm~\ref{alg:occupancy_grid_mapping}, performs a targeted subdivision at the leaf node to unlock navigable free space in OGM for motion planning. This process retains a significant portion of the free space that would otherwise be deemed untraversable. The resulting OGM thus provides} a substantially larger configuration space for path planners, directly enhancing both the probability of finding a valid path and the quality of the solution in complex environments.

\section{POMP Design and Implementation}
In this section, we present the implementation of the data structures and algorithms. The terminology and notation used in this paper are summarized in Table~\ref{tab:terminology}.

\subsection{Data Structure}

In \textbf{Data Structure 1}, Lines 2 and 3 define common attributes of a standard Octree node: the node’s size and pointers to its child nodes. The atomic type is used to enable concurrent (thread-safe) modifications to the child pointers without requiring a mutex, allowing safe node creation and point insertion in parallel. \yihui{Line 3 stores atomic child pointers, which enable concurrent child creation 
during parallel insertion. In 2D, each node has four children, whereas in 3D it has eight. In line 4, we optionally store point coordinates in each leaf node using a concurrent vector.} A concurrent vector is a thread-safe data structure that allows multiple threads to insert or access elements simultaneously without explicit locking, thus preventing race conditions during tree construction. Line 5 defines \texttt{safe\_state}: \yihui{an atomic bitmask records the occupancy and safety of regions within a leaf node (8 bits in 2D and 16 bits in 3D). Using an atomic type ensures thread-safe updates during concurrent point insertions.}

\renewcommand{\algorithmcfname}{Data Structure}
\renewcommand{\thealgocf}{1} 
\begin{algorithm}[h!]
\caption{OctreeNode }
\textbf{Structure:}  \\
\Indp 
    double \texttt{node\_size};  \\array$<$atomic$<$OctreeNode*$>$,8$>$ \texttt{children};\\
    concurrent\_vector$<$PointType$>$  \texttt{points}; \\
    atomic$<$uint16\_t$>$ \texttt{ safe\_state}\{0\};

\Indm
\end{algorithm}

\textbf{Data Structure 2} \yihui{includes} the pointer to the top-level node of the Octree, serving as the root \yihui{for} accessing and traversing the entire \yihui{tree} in Line 2. \yihui{ The  \texttt{map} in Line 4 is for pathfinding, which is formally introduced in \textbf{Data Structure 3}.}
The \texttt{leafsize} \yihui{is a given value that} represents the leaf node size, and is the same \yihui{as the resolution of the map in \textbf{Data Structure 3}.}

\renewcommand{\algorithmcfname}{Data Structure}
\renewcommand{\thealgocf}{2} 
\begin{algorithm}[h!]
\caption{Octree }
\textbf{Structure:}  \\
\Indp
    OctreeNode*  \texttt{ root};\\
    double \texttt{leafsize}; \\
    \xuehui{OccupancyGridMap*}  \texttt{map};\\
\Indm
\end{algorithm}

\textbf{Data Structure 3} is the \yihui{OccupancyGridMap}, \yihui{where} \yihui{Line 3} references an array of unsigned char (1 byte) occupancy state for thread-safe parallel updates. It plays the same role as \yihui{occupancy grid} cells \yihui{but is designed to} minimize conservative full-occupancy labels, improving spatial efficiency. The \texttt{resolution} \yihui{(Line 2)} \yihui{represents} the size of each voxel in the map.

\renewcommand{\algorithmcfname}{Data Structure}
\renewcommand{\thealgocf}{3} 
\begin{algorithm}[h!]
\caption{OccupancyGridMap }
\textbf{Structure:}  \\
\Indp
    double  \texttt{resolution};\\
    atomic$<$unsigned char$>$* \texttt{voxels};\\
\Indm
\end{algorithm}

\subsection{Parallel Octree Construction} \label{sec:octree construction}

\begin{figure}[h!]
    \centering
    \includegraphics[width=1.0\linewidth]{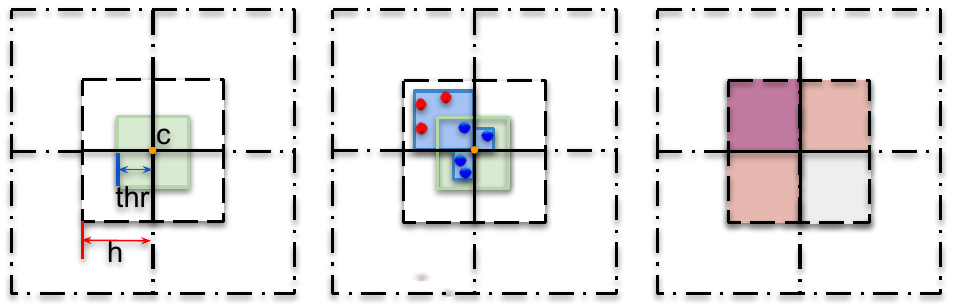}
    \caption{Illustration of the threshold setup and region classification: \texttt{clear}  ,  \texttt{safe},  and \texttt{unsafe}. The left panel depicts the boundary of the \colorcube{mysoftSage} area, where \(h\) denotes half the edge length of an OctoMap leaf node (equal to the map resolution r), and the threshold distance is \(\mathrm{thr}=h\cdot\mathrm{ratio}\), with \(\mathrm{ratio}\) specified at initialization. Red points lie outside the thresholded boundary, while blue points lie inside; these points determine the region safe state label (\texttt{unsafe} / \texttt{safe} / \texttt{clear}).
 }
    \label{fig:trd}
\end{figure}

\begin{figure*}[t!]
    \centering
    \includegraphics[width=0.98\linewidth]{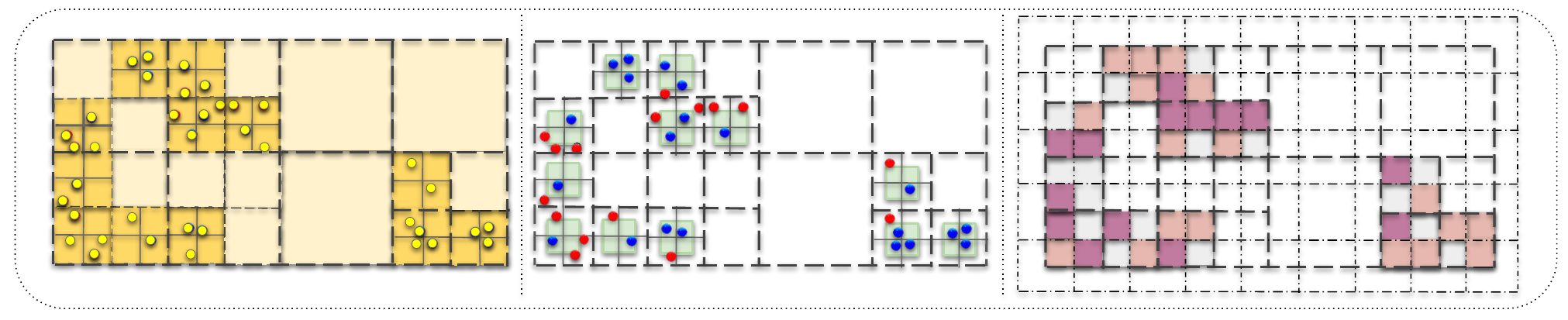}
    \caption{Illustration of the mapping configuration from Octree leaf nodes (left) to leaf regions and their projection onto the OGM (right). The occupancy grid is shown with a dash--dot outline, OctoMap nodes with a long-dashed outline, and region partitions with solid lines. }

    \label{fig:map_setup}
\end{figure*}

\renewcommand{\algorithmcfname}{Algorithm}
\renewcommand{\thealgocf}{1} 
\begin{algorithm}[t]
\caption{Parallel Octree Build}
\label{alg:parallel_octoMap_build}
\KwIn{Points $\mathcal{P}$}
\SetKwFunction{Fbuild}{ TreeBuild}
\SetKwProg{Fn}{Function}{:}{End Function}
\setcounter{AlgoLine}{0}
\Fn{\Fbuild{$\mathcal{P}$}}{

\ForEach{$p \in \mathcal{P}$ \textbf{in parallel}}{
     \texttt{ParallelInsertPoints}$(p, \textit{root})$
}

}

{\let\theAlgoLine\relax \rule{0.90\linewidth}{0.4pt}\\[1ex]}

\KwIn{Points p, OctreeNode node}
\SetKwFunction{FMain}{ ParallelInsertPoints}
\SetKwProg{Fn}{Function}{:}{End Function}
\setcounter{AlgoLine}{0}
\Fn{\FMain{$p$, \texttt{node}}}{
    \If{\texttt{node} is leaf}{
        push $p$ into \texttt{node.points}\\
        \texttt{set\_safe\_state}($p$, \texttt{node})\\
        \Return{}
    }
    \Else{
        $idx \leftarrow \texttt{node.get\_child\_idx}(p)$\\
        $\texttt{childPtr} \leftarrow \texttt{node.children}[idx]$\\
        $\texttt{child} \leftarrow \texttt{AtomicLoad}(\texttt{childPtr})$\\

        \If{\texttt{child} is \texttt{nullptr}}{
            $\texttt{newNode} \leftarrow$ new \texttt{OctreeNode}\\
            $\texttt{old} \leftarrow \texttt{nullptr}$\\
            $\texttt{ok}\leftarrow \texttt{CAS}(\texttt{childPtr},\texttt{old},\texttt{newNode})$\\

            \If{\texttt{ok}}{
                $\texttt{child} \leftarrow \texttt{newNode}$\\
            }
            \Else{
                \texttt{destroy(newNode)}\\
                $\texttt{child} \leftarrow \texttt{AtomicLoad}(\texttt{childPtr})$\\
            }
        }

        \texttt{ParallelInsertPoints}($p$, \texttt{child})\\
    }
}
    




{\let\theAlgoLine\relax \rule{0.90\linewidth}{0.4pt}\\[1ex]}

\KwIn{Points p, OctreeNode node}
\SetKwFunction{FMain}{ set\_safe\_state}
\SetKwProg{Fn}{Function}{:}{End Function}
\setcounter{AlgoLine}{0}
\Fn{\FMain{$p$, \texttt{node}}}{
 $c\leftarrow$ node.center; \\
 $h  \leftarrow$ node.size * 0.5; \\
 idx $\leftarrow$ node.\texttt{get\_child\_idx($p$)}\\
 offset $\leftarrow$ number\_of\_regions \\
mask $\leftarrow$ \texttt{1u<<(idx+offset)}\\
\If{$\lVert p - c\rVert_{\infty} \ge h \cdot \textit{ratio}$}
{ mask $\leftarrow$ \texttt{mask|1u<<(idx)}\\}
\texttt{AtomicOr(node.safe\_state,mask)}\\
}

\end{algorithm}

The parallel Octree construction follows the standard octree insertion procedure, which is illustrated in Fig. \ref{fig:octomap_build},  but inserts points concurrently using multiple threads (see \textbf{Algorithm 1}). \yihui{We parallelize point insertion by distributing the input points across threads; each thread traverses the tree from the root to the corresponding leaf for every point it inserts.} \yihui{Using parallel threads to access the same node can lead to race conditions; therefore, whenever constructing a new child under a parent node, we first check whether the pointer is nullptr.} \yihui{An atomic compare-and-swap (CAS) is used to safely construct child nodes during parallel point insertion. A new node is constructed only if the child pointer remains \texttt{nullptr} at the time of the operation.} \yihui{A successful CAS atomically sets the corresponding child pointer to the newly allocated node,} whereas failure indicates that another thread has already constructed the child, in which case the newly allocated node is discarded.  This process is repeated recursively until reaching a leaf node, whose size corresponds to the predefined resolution.

Each OctoMap leaf node is subdivided into four regions in 2D or eight regions in 3D. 
During the point insertion (see the \texttt{set\_safe\_state} function), 
each point is checked against the leaf node center $c$. \yihui{A region is marked \texttt{unsafe} if it contains any point whose perpendicular distance to \emph{any} axis-aligned splitting line/plane (used to partition the leaf cube) exceeds the threshold $\textit{thr}=h\cdot\textit{ratio}$, where $h$ is half of the leaf-cube edge length. Equivalently, $\max_i |p_i-c_i|\ge\textit{thr}$, where $p_i$ and $c_i$ are the $i$-th coordinate components of the point $p$ and the leaf center $c$, respectively, and the splitting planes pass through $c$.}

This threshold provides a margin so that points too close to the boundary are conservatively treated as obstacles. 
If all points in the region lie within this margin, the region is marked \texttt{safe}. 
If the region contains no points, it is marked \texttt{clear}.  All settings are illustrated in Fig.~\ref{fig:trd}. For compact storage, these states are encoded in a single \texttt{safe\_state} variable. In 2D, for each region $i$, bit $i+4$ records whether the region is non-clear, i.e., whether it contains at least one point, and bit $i$ records whether the
region is \texttt{unsafe}. Thus, a region is \texttt{clear} if neither bit is set, \texttt{safe} if only the non-clear bit is set, and \texttt{unsafe} if both the non-clear bit and the unsafe bit are set. The 3D case uses the same encoding with eight regions, where bits $8$--$15$ record non-clear states and bits
$0$--$7$ record unsafe states. Note that \texttt{unsafe}, \texttt{safe}, and \texttt{clear} are the three region states we use to evaluate safety for navigability. A region is non-clear if it has been visited by points, which includes both the \texttt{unsafe} and \texttt{safe} cases. This atomic encoding allows multiple threads to update the safe state of different regions in parallel without requiring locks. The state of each region determines the navigability of its corresponding occupancy grid/voxel.


\subsection{Mapping Implementation} \label{sec:mapping}

\begin{figure}[h!]
    \centering
    \includegraphics[width=1.0\linewidth]{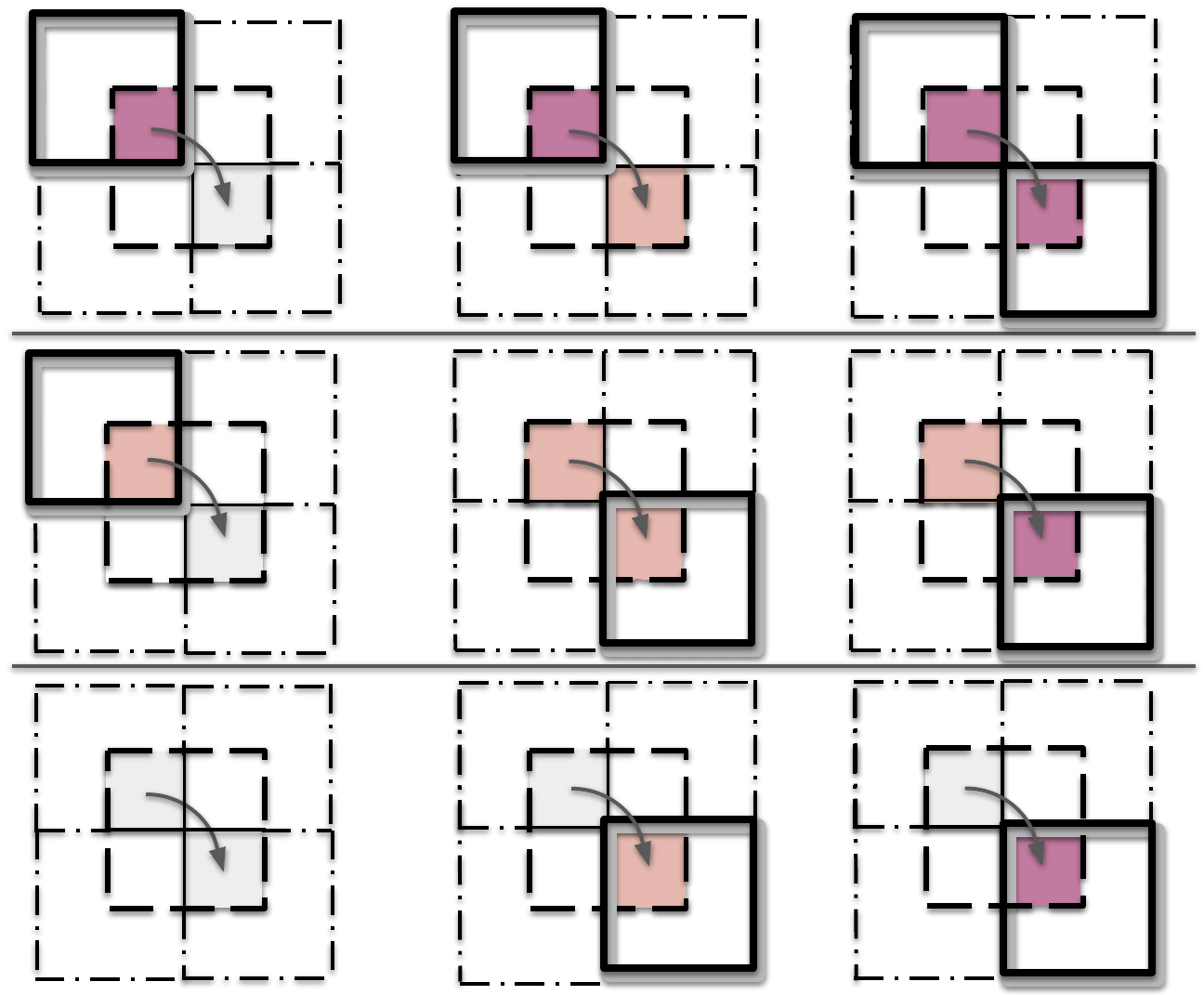}
    \caption{Illustration of the diagonal examination pairs and rules for determining grid navigability. \colorcube{myPink} is the \texttt{unsafe} region in the \yihui{leaf node}, \colorcube{softCoral} is the \texttt{safe} region in the leaf node, and \colorcube{lightGrayish} is the \texttt{clear} region in the leaf node. The arrow represents the direction of examination. The floating square with a black border indicates that the corresponding \yihui{cell of OGM} should be marked as \texttt{occupied} after the examination.
    }
    \label{fig:mapping_method}
\end{figure}

\renewcommand{\algorithmcfname}{Algorithm}
\renewcommand{\thealgocf}{2} 
\begin{algorithm}[t]
\caption{Occupancy Grid Mapping}
\label{alg:occupancy_grid_mapping}
\KwIn{OctreeNode node}
\SetKwFunction{FMain}{project\_onto\_OGM}
\SetKwProg{Fn}{Function}{:}{End Function}
\setcounter{AlgoLine}{0}
\Fn{\FMain{\texttt{node}}}{
  \If{region[from\_idx] is \texttt{clear}}{
     \If{region[to\_idx] is not \texttt{clear}}{
           \texttt{set\_cell\_Occupied(to\_idx)}
     }
  }
  \Else{
    \If{region[from\_idx] is \texttt{safe}}{
        \If{region[to\_idx] is not \texttt{clear}}{
            \texttt{set\_cell\_Occupied(to\_idx)}
        }
        \Else{
           \texttt{set\_cell\_Occupied(from\_idx)}
        }
    }
    \Else{
        \If{region[to\_idx] is \texttt{unsafe}}{
            \texttt{set\_cell\_Occupied(to\_idx)}
        }
        \texttt{set\_cell\_Occupied(from\_idx)}
    }
  }
}
{\let\theAlgoLine\relax \rule{0.90\linewidth}{0.4pt}\\[1ex]}
\KwIn{OctreeNode node}
\SetKwFunction{Fbuild}{traverse}
\SetKwProg{Fn}{Function}{:}{End Function}
\setcounter{AlgoLine}{0}
\Fn{\Fbuild{\texttt{node}}}{
  \If{node is Leaf}{
      \texttt{project\_onto\_OGM}$(\textit{node})$
  }
  \Else{
    \ForEach{$child \in node$ \textbf{in parallel}}{
        \texttt{traverse(child)}
    }
  }
}
\end{algorithm}

\begin{figure*}[t]
    \centering
    \includegraphics[width=1.0
    \linewidth]{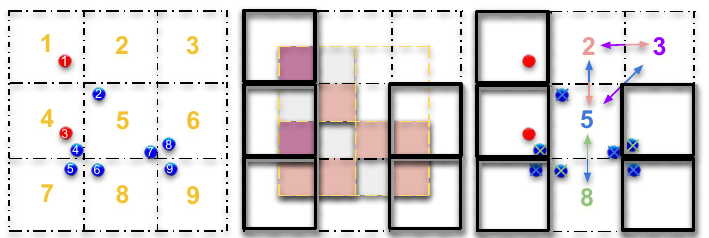}
\caption{\yihui{Illustration of the \xuehui {OGM} produced by our method. This panel shows the bottom-right corner of the map in Fig.~\ref{fig:map_setup}. For clarity, the OGM cell boundaries are drawn with a black dash-dotted outline, and the leaf-node boundary is shown as a gold long-dashed box. The nine OGM cells are labeled 1--9, and nine representative points are shown, colored by whether they lie inside or outside the threshold bound. The occupancy of diagonal pairs of OGM cells is determined by using Fig.~\ref{fig:mapping_method}. The arrows indicate the feasible navigations between grid cells, and the colors encode each cell’s feasible navigations. Colored crosses mark the point cloud blocked during the Diagonal Examination for the corresponding colored cells.}
}
\label{fig:ogm_nav}
\end{figure*}

The OctoMap and the \yihui{OGM} are tightly coupled: the OctoMap is continuously updated online, as described in Section~\ref{sec:octree construction}, and the OGM used for pathfinding is derived from the OctoMap. The \yihui{region safe states in the leaf level}, as shown in Fig.~\ref{fig:trd}, are used to infer the navigability of the corresponding OGM cells, and the inferred navigability is then used to assign the occupancy of each cell. Fig.~\ref{fig:map_setup} illustrates the spatial correspondence between the OctoMap and the OGM over the entire map.

\yihui{In Algorithm~\ref{alg:occupancy_grid_mapping}, the constructed OctoMap is traversed from the root to the leaves using the function \yihui{\texttt{traverse(node)}}. At each leaf node, we call \yihui{\texttt{project\_onto\_OGM(node)}} to mark the occupancy of the corresponding OGM cell. This function applies diagonal examination (see Fig.~\ref{fig:mapping_method}) to determine the navigability of the OGM cell containing the region, as well as the navigability of the OGM cell containing its diagonal region, and accordingly assigns the occupancy of that cell using \texttt{set\_cell\_Occupied(idx)}.} Here, \texttt{from\_idx} and \texttt{to\_idx} follow the arrow direction illustrated in Fig.~\ref{fig:mapping_method}.

\begin{table*}[t!]
\centering
\caption{Navigability State Determination Table based on the Diagonal Examination}
\renewcommand{\arraystretch}{1.25}
\setlength{\tabcolsep}{1.95pt}
\begin{tabular}{@{}>{\raggedright}p{0.11\linewidth} >{\raggedright}p{0.155\linewidth} >{\raggedright}p{0.155\linewidth} >{\raggedright}p{0.27\linewidth} >{\raggedright\arraybackslash}p{0.28\linewidth}@{}}
\toprule
\textbf{Region pair} & \textbf{Axis-dir. navigability} & \textbf{Diag-dir. navigability} & \textbf{Marking in cell} & \textbf{Notes} \\
\midrule
\textbf{[unsafe unsafe]} &
unsafe: \unsafeText{\ding{55}}\hspace{1ex} unsafe: \unsafeText{\ding{55}} &
unsafe: \unsafeText{\ding{55}}\hspace{1ex} unsafe: \unsafeText{\ding{55}} &
Marking both diagonal cells as \texttt{occupied} &
Axis-aligned: $\lVert p - c\rVert_{\infty} \ge h \cdot \textit{ratio}$ \(\Rightarrow\) \unsafeText{\ding{55}}
Diagonal: both regions contain points \(\Rightarrow\) \unsafeText{\ding{55}}
\\
\textbf{[safe safe]}  &
safe: \safeText{\ding{51}}\hspace{3.1ex} safe: \safeText{\ding{51}} &
safe: \safeText{\ding{55}}\hspace{3.1ex} safe: \safeText{\ding{55}} &
Mark one as \texttt{occupied} to block diagonal access while preserving axis-aligned navigation in the other region &
Axis-aligned: $\lVert p - c\rVert_{\infty} < h \cdot \textit{ratio}$ \(\Rightarrow\) \safeText{\ding{51}}
Diagonal: both regions contain points \(\Rightarrow\) \safeText{\ding{55}}
\\

\textbf{[clear clear]} &
clear: \clearText{\ding{51}}\hspace{2.1ex} clear: \clearText{\ding{51}} &
clear: \clearText{\ding{51}}\hspace{2.1ex} clear: \clearText{\ding{51}} &
Keep \texttt{unoccupied} in both diagonal cells &
Both regions do not contain points \(\Rightarrow\) \clearText{\ding{51}}  \\

\midrule
\textbf{[unsafe safe]} &
unsafe: \unsafeText{\ding{55}}\hspace{1ex} safe: \safeText{\ding{51}} &
unsafe: \unsafeText{\ding{55}}\hspace{1ex} safe: \safeText{\ding{55}} &
Mark the unsafe cell as \texttt{occupied}. Allow axis-aligned navigation in the safe region.  &
Axis-aligned:  $\lVert p - c\rVert_{\infty} \ge h \cdot \textit{ratio}$ \(\Rightarrow\) \unsafeText{\ding{55}}
\hspace*{5.6em}   $\lVert p - c\rVert_{\infty} < h \cdot \textit{ratio}$ \(\Rightarrow\) \safeText{\ding{51}}
Diagonal: both regions contain points \(\Rightarrow\) \unsafeText{\ding{55}}
\\
\textbf{[unsafe clear]} &
unsafe: \unsafeText{\ding{55}}\hspace{1ex} clear: \clearText{\ding{51}} &
unsafe: \unsafeText{\ding{55}}\hspace{1ex} clear: \clearText{\ding{51}} &
Mark the unsafe cell as \texttt{occupied}. Allow axis-aligned and diagonal navigation in the clear region.  &
Axis-aligned:  $\lVert p - c\rVert_{\infty} \ge h \cdot \textit{ratio}$ \(\Rightarrow\) \unsafeText{\ding{55}}
Diagonal: unsafe region is non-empty \(\Rightarrow\) \unsafeText{\ding{55}}
The clear region contains no points  \(\Rightarrow\) \clearText{\ding{51}}
\\

\textbf{[safe clear]} &
safe: \safeText{\ding{51}}\hspace{3.1ex} clear: \clearText{\ding{51}} &
safe: \safeText{\ding{55}}\hspace{3.1ex} clear: \clearText{\ding{51}} &
Mark the safe cell as \texttt{occupied}. Allow axis-aligned and diagonal navigation in the clear region.&
Axis-aligned: $\lVert p - c\rVert_{\infty} < h \cdot \textit{ratio}$\(\Rightarrow\) \safeText{\ding{51}}
Diagonal: safe region is non-empty \(\Rightarrow\) \safeText{\ding{55}}
The clear region contains no points  \(\Rightarrow\) \clearText{\ding{51}}
\\
\bottomrule
\end{tabular}\\
\vspace{4pt}
\parbox{\textwidth}{\scriptsize
\emph{Note:} \navYes/\navNo{} denote navigable/non-navigable in the indicated direction; color denotes the region state. Notes summarize the navigability criteria for each region pair, and the Marking in cell column describes the POMP cell-marking strategy for preserving navigability.
}
\label{tab:diagonal-rules}
\end{table*}

\subsection{Diagonal Examination } \label{sec:DiagonalExamination}

In a standard occupancy grid map, a free cell has up to $8$  neighbors in 2D and up to $26$ neighbors in 3D. Accordingly, motion primitives can be categorized into two types: \yihui{axis-aligned and diagonal moves}.

Under our alignment configuration, the OGM is shifted by a half-cell offset so that the two discretizations are staggered by half of the cell size. Consequently, each OGM cell overlaps multiple OctoMap leaf regions: four in 2D and eight in 3D. A \texttt{clear} region contains no obstacles and is navigable in \yihui{all} directions; a \texttt{safe} region is \yihui{assumed to be} navigable only in \yihui{axis-aligned} directions; and an \texttt{unsafe} region is \yihui{not navigable} in any direction. We leverage this decomposition and let the regions jointly determine whether \yihui{an} OGM cell should be marked as occupied (see Fig. \ref{fig:ogm_nav}).

By design, any cell that does not contain an unsafe region is navigable in axis-aligned directions. Thus, we only need to design occupancy rules using diagonal navigability. For example, all cells except for cells 1 and 4 in Fig.~\ref{fig:ogm_nav} are axis-aligned navigable. Cells 6, 7, and 9 are nevertheless marked as occupied due to diagonal navigability considerations.

Using the presence of points in leaf-node regions to directly determine diagonal navigability is overly conservative. For example, in Fig.~\ref{fig:ogm_nav}, although all cells except 2 and 3 appear to be diagonally non-navigable due to the presence of the points marked in blue, cells 5 and 8 can be safely marked unoccupied, which preserves axis-aligned navigation through cells 2, 5, and 8, and diagonal navigation between cells 3 and 5.

\yihui{Navigability of diagonal pairs of OGM cells is determined using the corresponding diagonal regions of the leaf node that overlaps them. In the 2D case, the four regions of a leaf node yield two diagonal pairs, labeled $[1,4], [2,3]$ in Fig.~\ref{fig:octomap_build}.
In the 3D case, the eight regions of a leaf node yield four diagonal pairs, labeled $[1,8], [2,7],[3,6],[4,5]$ in Fig.~\ref{fig:octomap_build}. As shown in Fig.~\ref{fig:mapping_method}, the navigability of diagonal regions is order-independent. Accordingly, in 2D only two diagonal pairs need to be examined, whereas in 3D only four diagonal examination directions are required.}

\begin{figure*}[t!]
    \centering
    \includegraphics[width=1.0\linewidth]{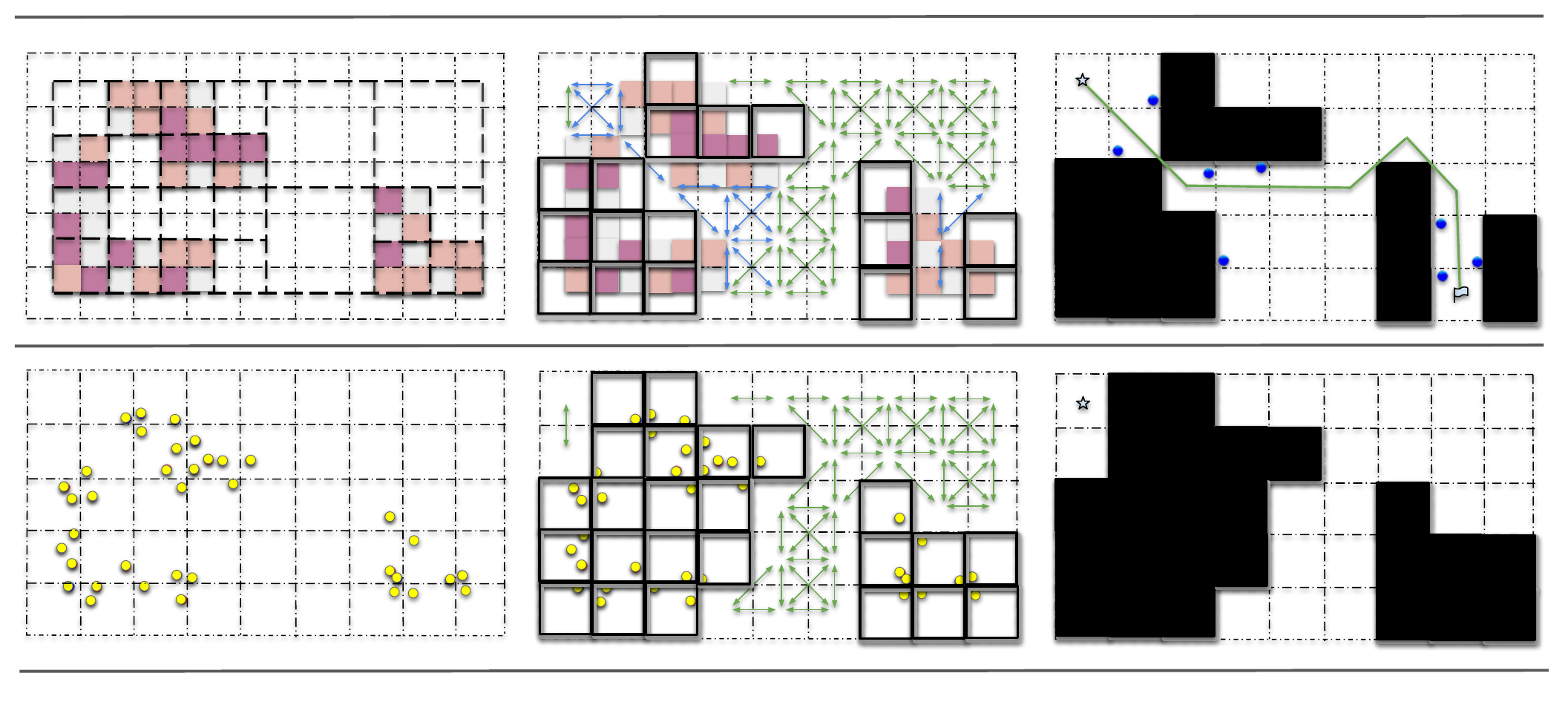}
    \caption{Top row (POMP): Using the obstacle setup in Fig.~\ref{fig:map_setup} and the mapping rule in Fig.~\ref{fig:mapping_method}, POMP assigns each region an \texttt{unsafe/safe/clear} label (top left), updates the corresponding occupancy grid cells and \yihui{their valid navigations between cells} (top middle), and produces the final occupancy map (top right). Compared to direct OGM, \yihui{POMP preserves more valid navigations (blue)}. Bottom row (direct OGM): For the same point cloud map, a standard occupancy grid voxelizes space (bottom left), marks a cell as occupied if it contains any points and \yihui{marks the valid navigations between cells} (bottom middle), and yields the final occupancy map (bottom right).}
    \label{fig:mapping_process}
\end{figure*}
\begin{figure*}[h!]
  \centering

  \begin{subfigure}[t]{0.326\linewidth}
    \centering
    \includegraphics[width=\linewidth]{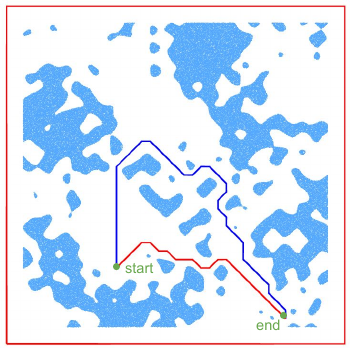}
  \end{subfigure}\hfill
  \begin{subfigure}[t]{0.326\linewidth}
    \centering
    \includegraphics[width=\linewidth]{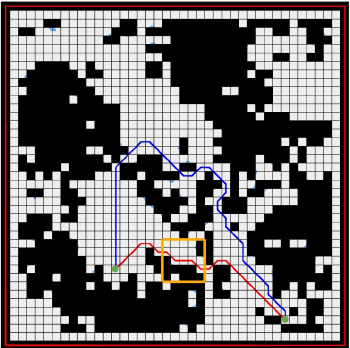}
  \end{subfigure}\hfill
  \begin{subfigure}[t]{0.326\linewidth}
    \centering
    \includegraphics[width=\linewidth]{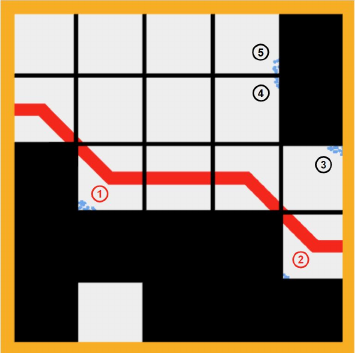}
  \end{subfigure}

\caption{\yihui{Paths planned using POMP (red) and direct (traditional) OGM (blue) superimposed on the point cloud (a) and on the occupancy grid (b). The orange box (enlarged on the right) highlights the region responsible for the shorter path. Differences between our POMP and direct OGM are highlighted in (c) by cell (1)-(5). These cells are classified as unoccupied by POMP but occupied by the direct OGM. In particular, keeping cells (2) and (3) free in our method explains the shorter path found, as it preserves a traversable narrow passage (see middle figure).
}
}

  \label{fig:grid_3x2}
\end{figure*}

\subsection{Navigability State Determination}  


\yihui{The rules for determining navigability of diagonal pairs of OGM cells are shown in Table~\ref{tab:diagonal-rules}. All cells in the OGM are initialized as \texttt{unoccupied}. Cells containing one or more unsafe regions are marked \texttt{occupied}. Cells containing only clear regions are marked \texttt{unoccupied}. Since safe regions are navigable in axis-aligned directions but not along diagonal directions, marking either the cell containing the safe region or its diagonally adjacent cell as \texttt{occupied} is sufficient to block diagonal navigation between the two cells (e.g., cells 1 and 5, cells 5 and 7, and cells 5 and 9 in Fig.~\ref{fig:ogm_nav}).}  

\yihui{Based on the above discussion, \yihui{the rules for occupancy assignment can be summarized as follows\xuehui{ (see Fig.~\ref{fig:mapping_method}):}}}

\begin{enumerate}
  \item \textbf{Safety hierarchy:} regions are prioritized in the order \yihui{of their safety levels} as \texttt{clear}, \texttt{safe}, and \texttt{unsafe}.
  \item \textbf{Mixed pair:} when two diagonally paired regions have different safety levels, the cell \yihui{containing the} safer region is designated as \texttt{unoccupied}, while the \yihui{cell containing the} less safe region is \texttt{occupied}.
  \item \textbf{Equal pair:} 
    \begin{itemize}
        \item if both regions are \texttt{unsafe}, both \yihui{cells} are \yihui{designated as} \texttt{occupied};
       \item if both regions are \texttt{safe}, designate either cell as \texttt{occupied} and the other as \texttt{unoccupied};
        \item if both regions are \texttt{clear}, \yihui{then both cells are designated as} \texttt{unoccupied};
    \end{itemize}
\end{enumerate}
 
\yihui{Note that each OGM cell overlaps multiple pairs of diagonal leaf-node regions (4 in 2D and 8 in 3D). If any of the diagonal pairs of leaf-node region results in an \texttt{occupied} designation, the cell is marked as occupied. Otherwise, it remains unoccupied (see Fig.~\ref{fig:mapping_process}). To prevent race conditions caused by cells participating in multiple examinations, we use atomic operations for both the region safe state update and the cell-level occupancy update. In addition, octree construction and projection onto the OGM are executed sequentially rather than asynchronously.}

\yihui{The comparison between POMP and direct OGM, using an identical point cloud, is shown in Fig.~\ref{fig:mapping_process} and Fig.~\ref{fig:grid_3x2}. In Fig.~\ref{fig:mapping_process}, the cells that contain points, but are marked by POMP as unoccupied, are highlighted in blue. These cells result in a feasible path connecting the start and the goal, whereas the direct OGM admits no such path. Fig.~\ref{fig:grid_3x2} shows another case where POMP produces a shorter (red) path by preserving more navigable space.}

\renewcommand{\algorithmcfname}{Algorithm}
\renewcommand{\thealgocf}{2}

\section{Experiments}
\yihui{In this section, we present randomized experiments to illustrate specific characteristics of the POMP algorithm, followed by offline experiments on real-world datasets and ROS bag simulations to demonstrate POMP’s performance. Except for Experiment B-1(a), all experiments are implemented in ROS 2 and conducted on an Intel Core Ultra 9 285 (2.50 GHz) system with 24 cores and 24 threads.}

\yihui{In the following,} five experiments on randomized datasets \yihui{are presented to illustrate the performance of POMP for} tree construction, \yihui{OGM construction, navigability, and planning.}

\subsection{Randomized \yihui{Point Clouds}}

\begin{figure}[h!]
    \centering
    \includegraphics[width=0.8\linewidth]{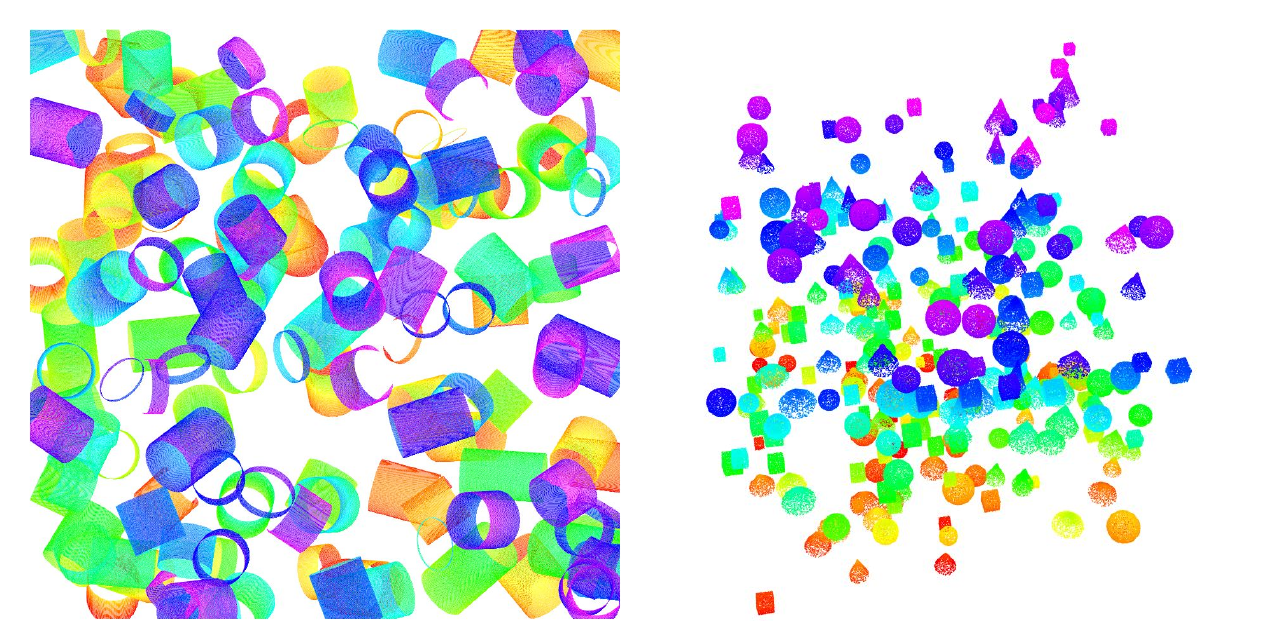}
    \caption{Two maps utilized in the randomized point-cloud experiments. }
    \label{fig:cylinder}
\end{figure}

\subsubsection{\yihui{Tree Construction}} In this experiment, we focus exclusively on evaluation of the computational efficiency \yihui{of the Octree construction module of POMP}.
\yihui{POMP is compared against} three state-of-the-art incremental tree construction methods: i-Octree, ikd-Tree, and PCL Octree, \yihui{for spatial representation} across different leaf sizes.
\begin{figure}[h!]
  \centering

  \begin{subfigure}{\linewidth}
    \centering
    \includegraphics[width=0.95\linewidth]{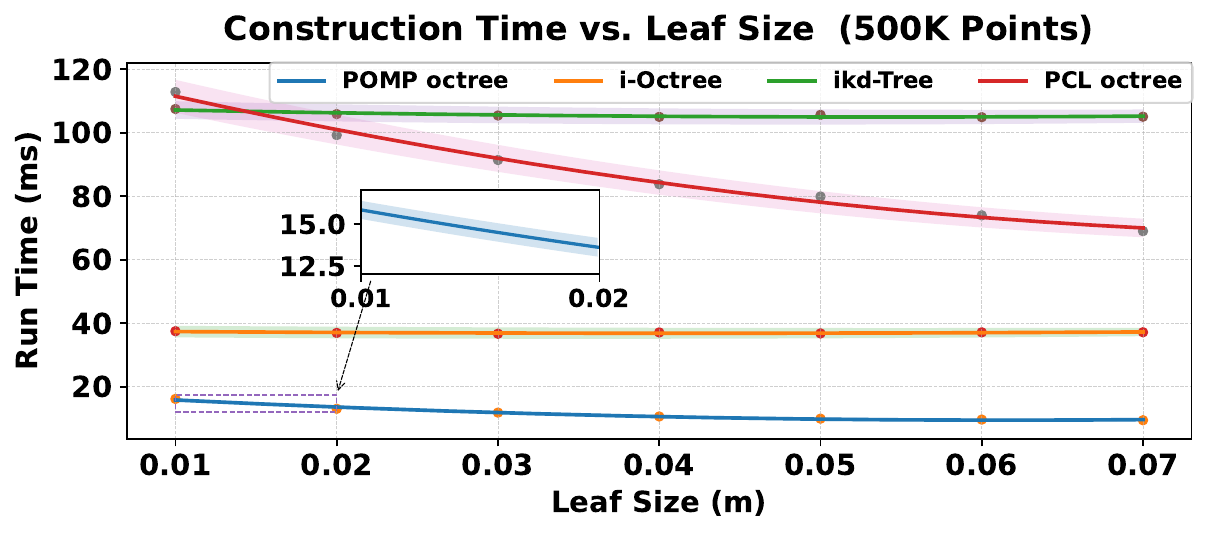}
  \end{subfigure}


  \begin{subfigure}{\linewidth}
    \centering
    \includegraphics[width=0.95\linewidth]{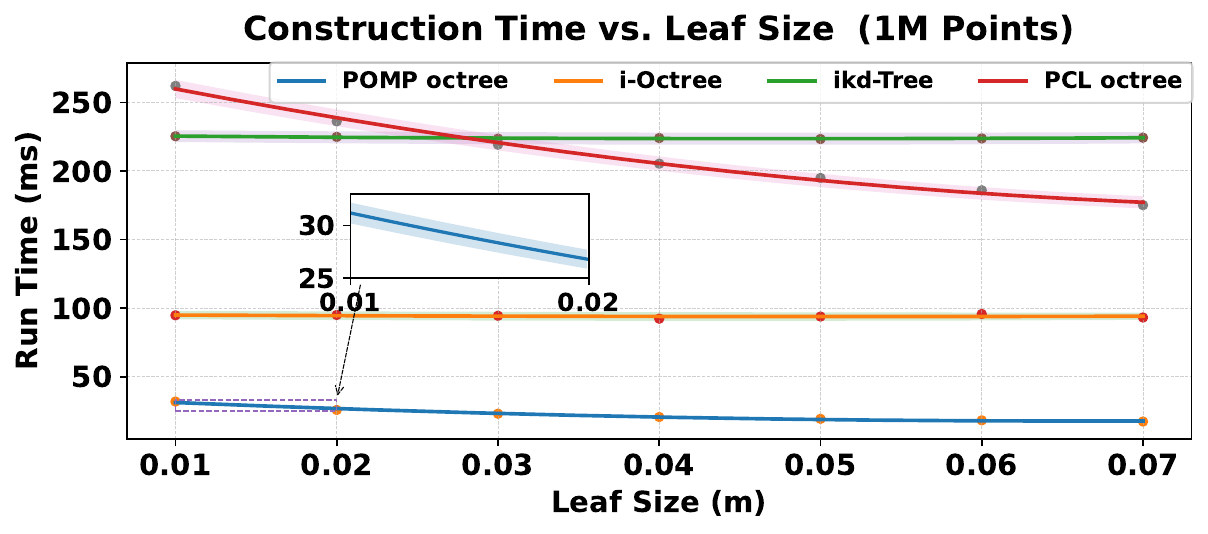}
  \end{subfigure}


  \begin{subfigure}{\linewidth}
    \centering
    \includegraphics[width=0.95\linewidth]{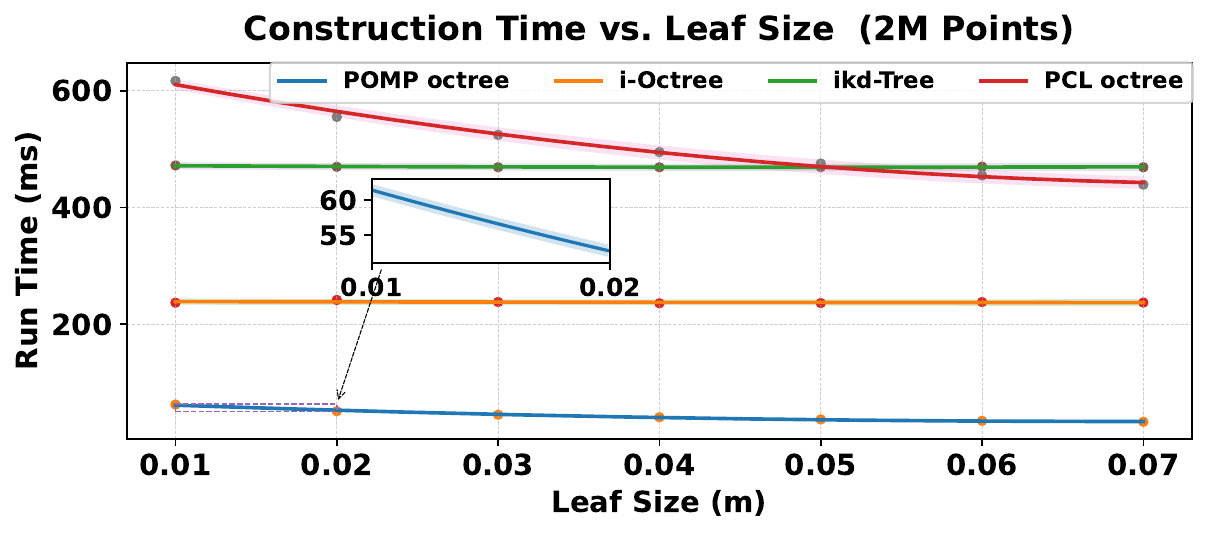}
  \end{subfigure}

  \caption{\yihui{\textit{Experiment A-1:}} Runtime comparison with \yihui{ point clouds, showing} 
    from top to bottom: $500{,}000$, $1{,}000{,}000$, and $2{,}000{,}000$ points.}
  \label{fig:Octree_construct_random}
\end{figure}
For evaluation, \yihui{datasets containing} \(500{,}000\), \(1{,}000{,}000\), and \(2{,}000{,}000\) points \yihui{are} randomly generated within a \(10 \,\text{m} \times 10 \,\text{m} \times 10 \,\text{m}\) workspace, and tree construction was repeated 200 times for each dataset \yihui{using} varying leaf sizes (0.01 m, 0.02 m, 0.03 m, 0.04 m, 0.05 m, 0.06 m and 0.07 m).








As shown in Fig. \ref{fig:Octree_construct_random}, across three datasets, POMP is at least 2 $\times$ faster than i-Octree \cite{SCC.Zhu.Li.ea2024}, 7–10 $\times$ faster than ikd-Tree\cite{SCC.Cai.Xu.ea2021}, and about 9 $\times$ faster than PCL Octree\cite{SCC.Rusu.Cousins.ea2011}, while maintaining similarly low standard deviations across resolutions. These results demonstrate that, from a data-structure perspective, POMP achieves consistently higher tree construction throughput than state-of-the-art tree \yihui{construction techniques}.


\subsubsection{OctoMap Construction \& Occupancy Mapping}

\yihui{The second experiment evaluates the total runtime of OctoMap construction and the subsequent projection of region safe states from leaf nodes onto OGM cell occupancy, and compares it with three baselines: a direct OGM method, a mutex-protected parallel OctoMap (OctoMap-MTX), and the original OctoMap.}
 In direct OGM method, the environment is discretized into grid cells, and each point in the point cloud directly updates the occupancy state of its corresponding cell. 
The OctoMap-MTX is a mutex-protected shared octree for parallel octree construction. To establish a strong lock-based baseline, we go beyond a naive shared-octree design with per-node mutexes. OctoMap-MTX still inserts all points into a single shared tree, but first partitions large point clouds by upper-level octree buckets and schedules the resulting disjoint subtrees to different workers. This retains the synchronization overhead of a shared-tree implementation while avoiding excessive fine-grained lock contention during bulk insertion. The original OctoMap baseline directly uses the official OctoMap implementation~\cite{hornung2013octomap}, where each input point updates the occupancy state of its corresponding octree node.

\yihui{This experiment} utilizes static uniform random cylinder environments. Cylinders are randomly distributed within the workspace defined by
$x \in [-10, 10]$, $y \in [-10, 10]$, and $z \in [-5, 5]$~(meters).
A total of 50 million, 5 million, and 500 thousand points are sampled, with the cylinder radius uniformly drawn from \([0.8, 1.0]\) m. To ensure robustness, 500 independent trial maps are generated under this configuration.

\begin{figure}[h!]
    \centering
    \includegraphics[width=1.0\linewidth]{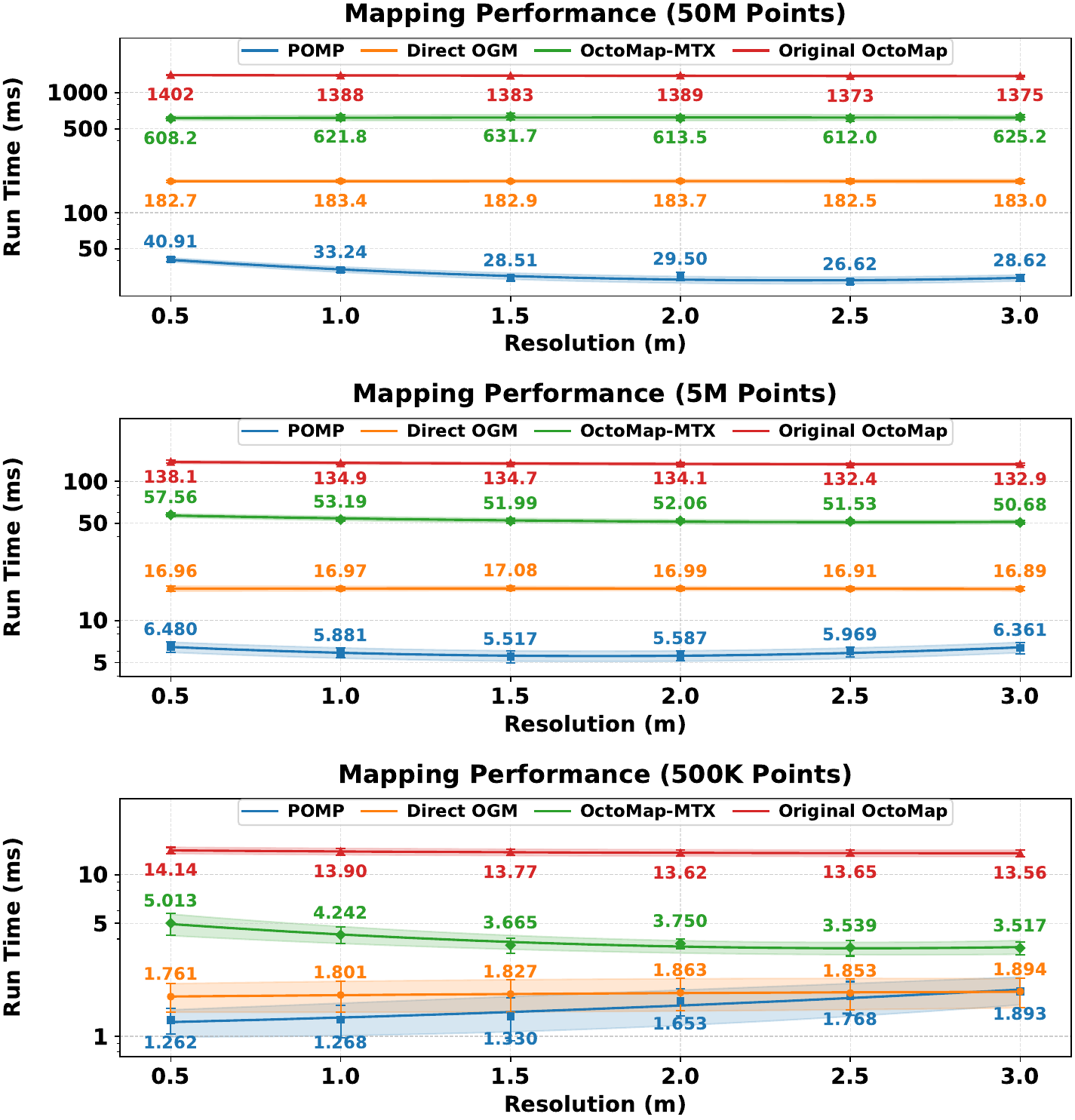}
    \caption{\yihui{Experiment A-2: Comparison of the total runtime of OctoMap construction and the subsequent projection of region safe states from leaf nodes onto OGM cell occupancy, against three baseline map representations (direct OGM, OctoMap-MTX and the original OctoMap). The y-axis is shown in log scale.}}
    \label{fig:mapping performance random}
\end{figure}
The results are shown in Fig. \ref{fig:mapping performance random}. To highlight the efficiency of our method, we compare our method with direct OGM. Across varying resolutions, our method is approximately \(4\text{--}6 \times\) faster than the \yihui{direct OGM method} for 50M points, \(2\text{--}3 \times\) faster for 5M points, and slightly faster for 500K points.  Notably, this runtime includes both OctoMap construction and the subsequent projection onto an OGM, but it still outperforms the direct method. These results indicate that POMP better supports real-time updates and continuous map availability, especially for dense point clouds, which is critical for online planning in large-scale, high-rate sensing scenarios.

\subsubsection{Navigability}

In the third experiment, we evaluate \yihui{the} navigability performance \yihui{of POMP} on randomized 3D worlds of size \(50\,\mathrm{m}\times50\,\mathrm{m}\times50\,\mathrm{m}\) uniformly populated with three geometric shapes, including spheres, right circular cones, and axis-aligned boxes. \yihui{The metric used to assess navigability is navigable space ratio (NSR), defined as the ratio of navigable cells to total cells}. To assess performance across spatial resolutions, we vary the resolution from \(0.5\) to \(5.0\,\mathrm{m}\) in \(0.5\,\mathrm{m}\) increments and generate \(200\) independent worlds for each resolution. Each world contains \(100\) spheres, \(100\) cones, and \(100\) boxes (300 obstacles in total), with their surfaces uniformly sampled into \yihui{generated point clouds containing} \(600{,}000\) points.

\begin{figure}[h!]
    \centering
    \includegraphics[width=1.0\linewidth]{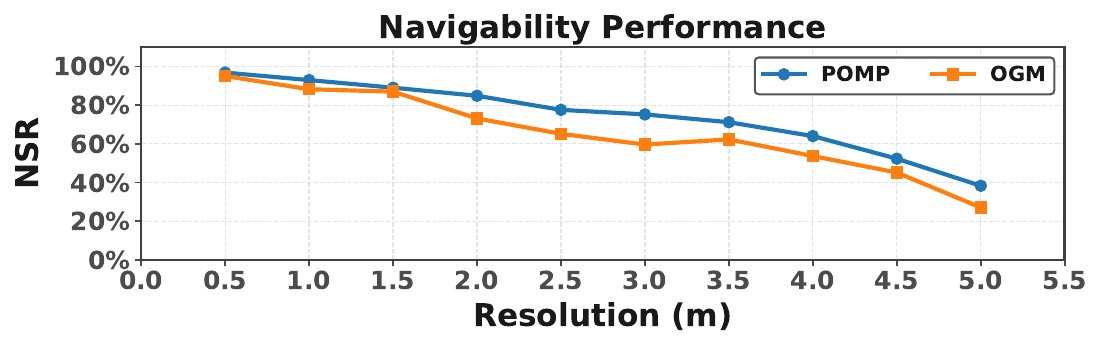}
    \caption{\yihui{Experiment A-3:} Comparison of the navigable space ratio (NSR) across \yihui{OGM cell resolutions.} }
    \label{fig:naviability_random}
\end{figure}

As shown in Fig. \ref{fig:naviability_random}, across varying resolutions POMP outperforms the \yihui{direct} occupancy grid \yihui{construction}, achieving up to about 10\% higher NSR at coarse resolutions. These results demonstrate that, at a fixed occupancy grid resolution, POMP refines the free space representation by reducing overly conservative occupancy labeling, thereby enlarging the feasible planning space for search-based planners.

\subsubsection{Planning Performance}

\begin{figure}[h!]
    \centering
    \includegraphics[width=1.0\linewidth]{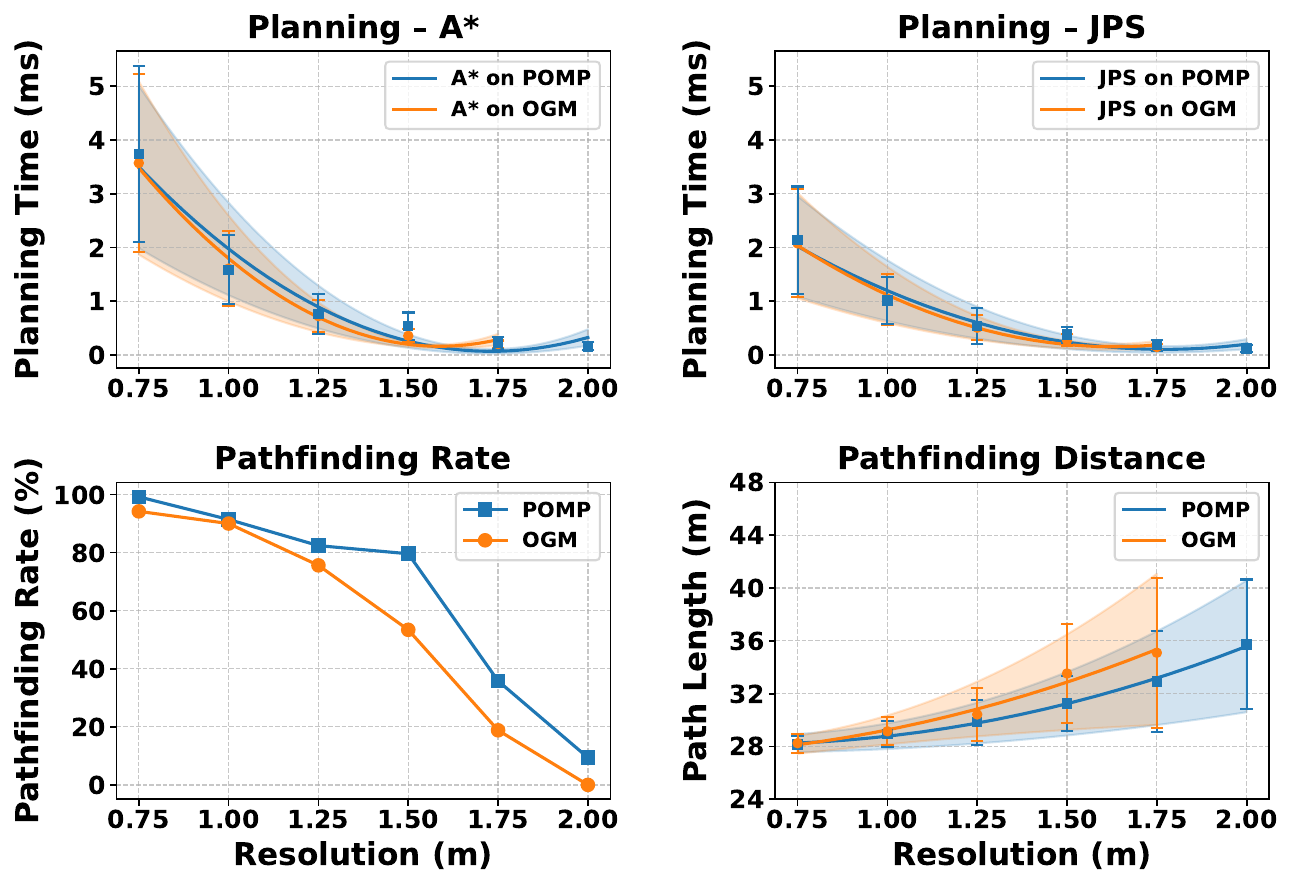}
    \caption{\yihui{Experiment A-4}: \yihui{Planning runtime and path length are reported} only when both approaches successfully found a path. Note that at a resolution of 2.0 m, across all 500 trial maps, no path was found in the direct OGM, but paths were still found in the POMP maps.}
    \label{fig:planning perforance random}
\end{figure}

The fourth experiment evaluates planning performance using the same randomized datasets as in the second experiment. The start and goal positions are fixed at \((-9, -9, -2)\) and \((9, 9, 2)\), respectively. A total of 500 independent trial maps are generated, and we evaluate the runtime of planning with two search-based pathfinding algorithms, A* and Jump Point Search (JPS). The evaluation metrics include planning runtime, pathfinding success rate, and path length, under both \yihui{direct} occupancy grid \yihui{construction} and \yihui{POMP} in varying resolutions. For a fair comparison, all cells outside the environment boundary \yihui{are marked} as occupied in all trials. \yihui{When comparing path lengths}, comparisons are made only when both approaches are able to find a path.

The results \yihui{in Fig. \ref{fig:planning perforance random}} indicate that, for search-based planners (A* and JPS), POMP incurs essentially no additional planning-time overhead compared with \yihui{direct OGM}. However, across varying map resolutions, POMP substantially improves the pathfinding success rate and consistently reduces the resulting path length.

\subsubsection{Performance Comparison Across Threshold Ratio}The fifth experiment evaluates the effect of the threshold ratio on planning performance. In contrast to the fourth experiment, we consider a continuous-motion scenario with a single environment populated by uniformly random cubes. Cubes are initially distributed within the workspace \(x \in [-25, 25]\), \(y \in [-25, 25]\), and \(z \in [-25, 25]\). \yihui{A total of 800 cubes are placed in the workspace, resulting in a point cloud of approximately 70{,}000 points. The cube side lengths are uniformly distributed over $[1.0,\,2.0]$~m.} Each cube independently selects a fixed random direction and then moves for 500 frames, translating \(1\) to \(2\ \mathrm{m}\) per frame. When a cube contacts the workspace boundary, its motion follows a specular reflection rule, with the angle of incidence equal to the angle of reflection. 

\yihui{Threshold} ratios of \(\textit{ratio} \in \{0.95, 0.75, 0.50, 0.25\}\) \yihui{are evaluated and compared against a direct OGM baseline}. \yihui{For path planning, the start and the goal} are fixed at \((-20,-20,-20)\) and \((20,20,20)\), respectively.

\begin{table}[h!]
  \centering
  \caption{\yihui{Experiment A-5:} Pathfinding Rate across 500 frames with varying threshold ratio}
  \label{tab:pomp_ogm_ratio}
  \setlength{\tabcolsep}{4pt}
  \renewcommand{\arraystretch}{1.05}
  \begin{tabular}{lcccccccc}
    \toprule
    & \multicolumn{7}{c}{\textbf{Resolution (m)}} \\
    \cmidrule(lr){2-8}
    & \textbf{1.0} & \textbf{1.5} & \textbf{2.0} & \textbf{2.5} & \textbf{3.0} & \textbf{3.5} & \textbf{4.0} \\
    \midrule
    \textbf{POMP 0.95} & 82.2\% & 68.8\% & 58.8\% &  50.0\% & 38.4\% & 23.0\%  & 15.4\% \\
    \textbf{POMP 0.75} & 82.2\% & 67.4\% &  57.7\%  & 46.4\% & 37.6\% & 21.8\%  & 14.4\% \\
    \textbf{POMP 0.50} & 81.4\% & 66.8\% &  56.3\%  & 43.0\% & 35.0\% & 19.6\% &  13.0\% \\
    \textbf{POMP 0.25} & 81.4\% & 66.4\% & 55.8\%  & 42.4\% & 34.0\% & 19.0\% & 12.4\% \\
    \textbf{OGM}       & 81.4\% & 66.4\%  & 55.6\%  & 42.0\% & 32.2\% & 16.2\%  & 9.6\% \\
    \bottomrule
  \end{tabular}
\end{table}

As reported in Table~\ref{tab:pomp_ogm_ratio}, decreasing the threshold ratio makes the safety margin more conservative, which reduces the effective navigable space and leads to a gradual decrease in planning success rate. \yihui{However}, for every evaluated resolution and ratio, POMP still outperforms the direct OGM baseline in pathfinding rate, demonstrating that \yihui{the} free-space refinement \yihui{feature of POMP} enlarges the feasible planning space while allowing the conservativeness to be tuned via \textit{ratio}.

\subsection{\yihui{Offline experiments on Real-world Datasets}}
\begin{figure}[h!]
    \centering
    \includegraphics[width=1.0\linewidth]{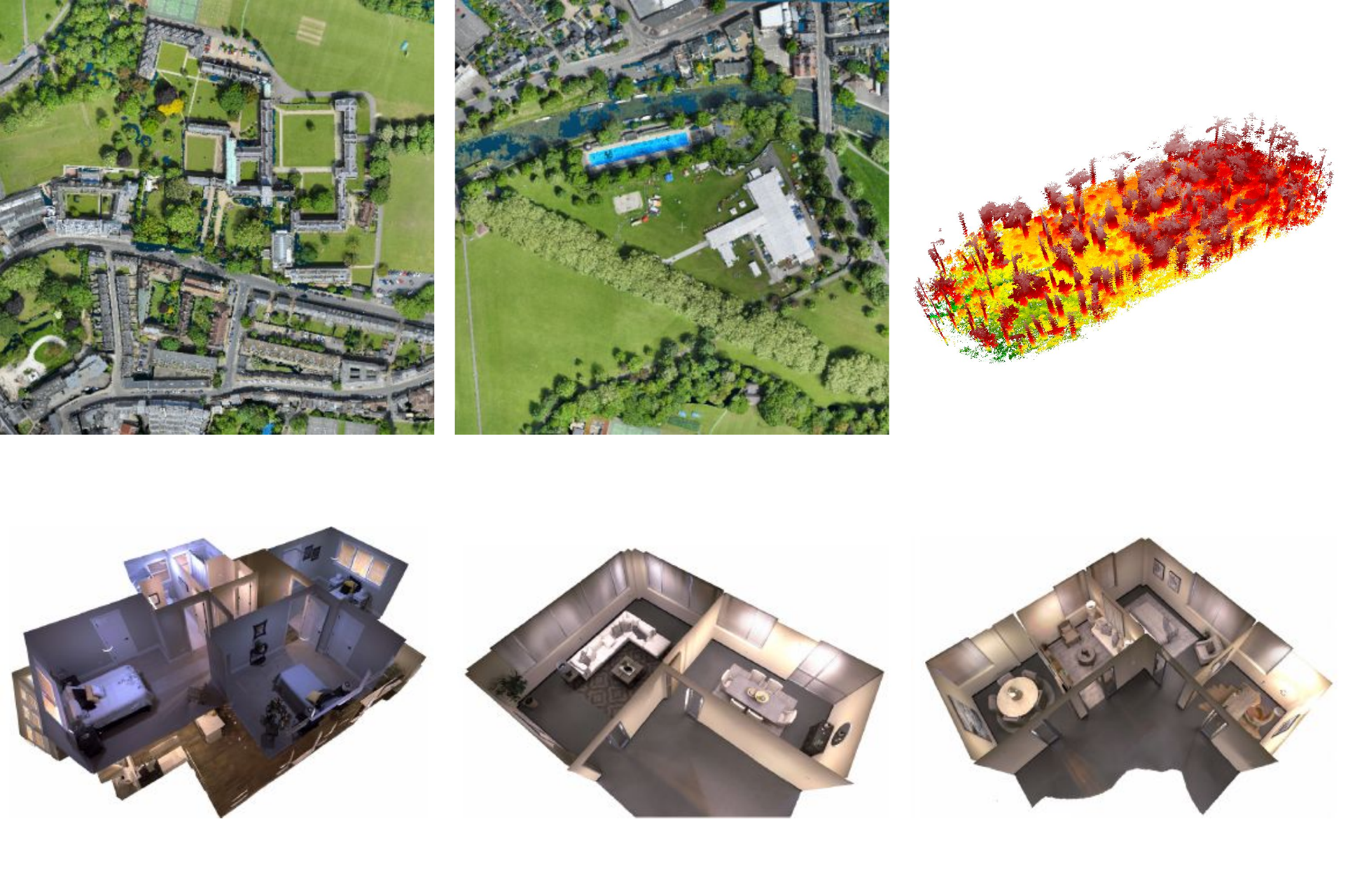}
    \caption{Benchmark maps used in the application experiments include Cambridge\_15, Cambridge\_16, Cloud\_7, Apartment\_0, Apartment\_1, and Apartment\_2.}
    \label{fig:six-wide}
\end{figure}
\begin{table}[h!]
\centering
\footnotesize
\setlength{\tabcolsep}{12pt}
\renewcommand{\arraystretch}{1.1}
\caption{Details of Maps in Datasets for Benchmark Experiment}
\label{tab:datasets}
\begin{tabular}{l c c}
\toprule
\textbf{Dataset} & \textbf{Map dimensions ($\text{m}^3$)} & \textbf{Point number} \\
\midrule
Cambridge\_15   & 400 $\times$ 400 $\times$ 60   & 119684095 \\
Cambridge\_16   &  400 $\times$ 400 $\times$ 35   & 113159239  \\
Cloud\_7         &  45 $\times$ 104 $\times$ 12   &12915992 \\
Apartment\_0    &   9.4 $\times$ 14.8 $\times$ 5.3      & 4564613  \\
Apartment\_1    & 10.7 $\times$ 7.9 $\times$ 2.9      & 1909995  \\
Apartment\_2    &  9.4 $\times$ 10.2 $\times$ 2.8      & 2136963  \\

\bottomrule
\end{tabular}
\end{table}

\begin{table*}[t!]
\centering
\footnotesize
\setlength{\tabcolsep}{3pt}
\renewcommand{\arraystretch}{1.1}
\caption{\yihui{Experiment B-1(a)}: Construction Runtime (ms) of OctoMap, Occupancy Grid Map, and Ours under Varying Resolutions on Five Maps (20-thread CPU)}
\label{tab:map_construction_labtop}
\begin{tabular}{l ccc ccc ccc ccc ccc}
\toprule
\textbf{Dataset} 
& \multicolumn{3}{c}{\textbf{Cambridge\_15}} 
& \multicolumn{3}{c}{\textbf{Cambridge\_16}} 
& \multicolumn{3}{c}{\textbf{Apartment\_0}} 
& \multicolumn{3}{c}{\textbf{Apartment\_1}} 
& \multicolumn{3}{c}{\textbf{Apartment\_2}} \\
\cmidrule(lr){2-4}\cmidrule(lr){5-7}\cmidrule(lr){8-10}\cmidrule(lr){11-13}\cmidrule(lr){14-16}
\textbf{Resolution (m)} 
& 1.0 & 3.0 & 5.0 
& 1.0 & 3.0 & 5.0 
& 0.05 & 0.10 & 0.50 
& 0.05 & 0.10 & 0.50 
& 0.05 & 0.10 & 0.50 \\
\midrule
\textbf{OctoMap} 
& 5089.00 & 4893.54 & 4878.26
& 4825.89 & 4626.40 & 4613.28
& 364.46  & 236.26  & 190.91
& 152.30  & 97.51   & 79.40
& 168.87  & 109.26  & 89.23 \\

\textbf{OctoMap-MTX}
& 2695.07 & 2650.77 &2511.81
& 2528.09 & 2446.34 & 2346.05
& 112.45 & 102.49 & 86.37
& 40.87 & 36.98 & 33.38
& 45.62 & 41.39 &  37.37 \\

\textbf{OGM}     
& 454.99 & 450.45 & 447.62
& 430.51 & 425.16 & 421.18
& \textbf{17.11}  & 17.68  & 16.94
& \textbf{7.16}   & 7.01   & 7.02
& 8.35   & 7.99   & 7.26 \\
\textbf{Ours}    
& \textbf{314.25} & \textbf{244.38} & \textbf{220.13}
& \textbf{296.04} & \textbf{233.06} & \textbf{214.62}
& 19.71  & \textbf{13.04}  & \textbf{8.07}
& 7.22   & \textbf{4.57}   & \textbf{4.04}
& \textbf{8.40}   & \textbf{5.40}   & \textbf{4.11} \\
\bottomrule

\end{tabular}

\end{table*}

\begin{table*}[t!]
\centering
\footnotesize
\setlength{\tabcolsep}{3.5pt}
\renewcommand{\arraystretch}{1.1}
\caption{\yihui{Experiment B-1(b)}: Construction Runtime (ms) of OctoMap, Occupancy Grid Map, and Ours under Varying Resolutions on Five Maps (24-thread CPU)}
\label{tab:map_construction_desktop}
\begin{tabular}{l ccc ccc ccc ccc ccc}
\toprule
\textbf{Dataset} 
& \multicolumn{3}{c}{\textbf{Cambridge\_15}} 
& \multicolumn{3}{c}{\textbf{Cambridge\_16}} 
& \multicolumn{3}{c}{\textbf{Apartment\_0}} 
& \multicolumn{3}{c}{\textbf{Apartment\_1}} 
& \multicolumn{3}{c}{\textbf{Apartment\_2}} \\
\cmidrule(lr){2-4}\cmidrule(lr){5-7}\cmidrule(lr){8-10}\cmidrule(lr){11-13}\cmidrule(lr){14-16}
\textbf{Resolution (m)} 
& 1.0 & 3.0 & 5.0 
& 1.0 & 3.0 & 5.0 
& 0.05 & 0.10 & 0.50 
& 0.05 & 0.10 & 0.50 
& 0.05 & 0.10 & 0.50 \\
\midrule
\textbf{OctoMap} 
& 3336.56 & 3222.84 & 3209.94
& 3167.78 & 3047.52 & 3039.78
& 220.76  & 144.46  & 121.60
& 88.18   & 59.50   & 51.10
& 99.44   & 67.50   & 57.98 \\
\textbf{OctoMap-MTX}
& 1873.78 &  1826.14 & 1721.92
& 1724.86 & 1677.24 & 1613.94
& 63.84 & 57.66 & 48.41
& 16.7 & 15.44 & 14.16
& 18.92 & 17.06 &  15.82 \\
\textbf{OGM}     
& 437.32 & 433.98 & 427.26
& 417.36 & 407.78 & 403.34
& 17.46  & 16.82  & 16.12
& \textbf{7.24} & 6.90 & 6.92
& \textbf{8.08} & 8.02 & 8.10 \\
\textbf{Ours}    
& \textbf{213.78} & \textbf{158.74} & \textbf{135.08}
& \textbf{205.08} & \textbf{151.14} & \textbf{128.54}
& \textbf{17.20}  & \textbf{11.78}  & \textbf{6.58}
& 8.02 & \textbf{5.88} & \textbf{3.16}
& 8.68 & \textbf{6.84} & \textbf{3.64} \\
\bottomrule
\end{tabular}
\end{table*}

\begin{table*}[t!]
\centering
\footnotesize
\setlength{\tabcolsep}{6.0pt}
\renewcommand{\arraystretch}{1.15}
\setlength{\aboverulesep}{0.25ex}
\setlength{\belowrulesep}{0.25ex}
\caption{\yihui{Experiment B-2 and B-3:} Pathfinding metrics across resolutions.}
\label{tab:planning_metrics_all}
\begin{tabular}{@{} l *{4}{c} *{4}{c} *{5}{c} @{}}
\toprule
& \multicolumn{4}{c}{\textbf{Cambridge\_15}}
& \multicolumn{4}{c}{\textbf{Cambridge\_16}}
& \multicolumn{5}{c}{\textbf{Cloud~7}} \\
\cmidrule(lr){2-5}\cmidrule(lr){6-9}\cmidrule(lr){10-14}
\textbf{Resolution (m)}
& 2 & 6 & 10 & 14
& 2 & 6 & 10 & 14
& 0.5 & 1.0 & 1.5 & 2.0 & 2.5 \\
\midrule
\textbf{OGM Pathfinding (\%)}
& 84.1 & 46.7 & 36.7 & 20.2
& 81.4 & 24.6 & 7.0 & 0.4
& 82.8 & 45.2 & 25.0 & 5.4 & 0.7 \\
\textbf{POMP Pathfinding (\%)}
& \textbf{91.5} & \textbf{68.8} & \textbf{50.2} & \textbf{45.1}
& \textbf{82.7} & \textbf{29.6} & \textbf{7.5} & \textbf{6.3}
& \textbf{92.1} & \textbf{68.4} & \textbf{44.9} & \textbf{33.6} & \textbf{16.4} \\
\midrule
\textbf{OGM Path Length (m)}
& 225.46 & 218.34 & 233.43 & 211.10
& 222.58 & 241.99 & 177.16 & -
& 45.90 & 48.08 & 52.62 & 33.23 & - \\
\textbf{POMP Path Length (m)}
& \textbf{225.42} & \textbf{218.30} & \textbf{233.40} & \textbf{210.90}
& \textbf{222.42} & \textbf{237.70} & \textbf{176.64} & -
& \textbf{45.84} & \textbf{47.70} & \textbf{50.94} & \textbf{31.22} & - \\
\midrule
\textbf{A* on OGM (ms)}
& 217.113 & \textbf{5.148} & \textbf{0.853} & \textbf{0.03}
& 130.048 & 3.045 & 0.071 & -
& \textbf{86.984} & \textbf{8.314} & \textbf{1.928} & \textbf{0.074} & - \\
\textbf{A* on POMP (ms)}
& \textbf{215.331} & 5.497 & 0.995 & 0.084
& \textbf{122.764} & \textbf{2.72} & \textbf{0.029} & -
& 91.589 & 10.246 & 2.692 & 0.167 & - \\
\midrule
\textbf{JPS on OGM (ms)}
& \textbf{114.408} & \textbf{1.518} & 0.204 & \textbf{0.001}
& 67.955 & 1.289 & 0.014 & -
& \textbf{43.827} & \textbf{4.454} & \textbf{1.104} & \textbf{0.019} & - \\
\textbf{JPS on POMP (ms)}
& 123.703 & 1.715 & \textbf{0.188} & 0.001
& \textbf{67.514} & \textbf{1.13} & \textbf{0.001} & -
& 48.234 & 5.239 & 1.524 & 0.074 & - \\
\bottomrule
\end{tabular}

\vspace{4pt}
\parbox{\textwidth}{\scriptsize
\emph{Note:} “-” indicates that insufficient common successful trials were available for the comparison of path length and pathfinding time.
}
\end{table*}
We evaluate \yihui{POMP in real-world scenarios using} three public datasets: \texttt{SensatUrban}~\cite{hu2022sensaturban}, \texttt{Replica}~\cite{replica19arxiv} and \texttt{Treescope}\cite{cheng2023treescope}. \texttt{SensatUrban} is an urban-scale photogrammetric point cloud dataset with nearly three billion points from three UK cities; Replica provides high quality reconstructions of diverse indoor environments. Treescope is the first robotics dataset in precision agriculture and forestry designed specifically for counting and mapping trees in forest and orchard environments. The datasets \texttt{Cambridge\_15} and \texttt{Cambridge\_16} from SensatUrban, \texttt{Apartment\_0}, \texttt{Apartment\_1}, and \texttt{Apartment\_2} from Replica, and \texttt{Cloud\_7}  from WSF-19 in Treescope \yihui{are used for evaluation} (See Fig. \ref{fig:six-wide}). Dataset details are provided in Table~\ref{tab:datasets}.

\subsubsection{Map Construction \yihui{Performance}}\yihui{Map construction} runtime \yihui{of POMP is evaluated} against three baselines: \yihui{direct OGM} and OctoMap-MTX with the same settings as in the randomized data experiment, and the original OctoMap, which updates occupancy via point cloud insertion. For OctoMap, we used the official open-source implementation available on GitHub~\cite{hornung2013octomap}.

\yihui{Since the} performance of POMP scales with thread count, we evaluated it on two machines: (i) an Intel Core i9-13900H system with 14 cores and 20 threads, and (ii) an Intel Core Ultra 9 285 (2.50 GHz) system with 24 cores and 24 threads. \yihui{As} shown in Tables \ref{tab:map_construction_labtop} and \ref{tab:map_construction_desktop}, the reported time covers the entire pre-planning pipeline, including \yihui{Octree construction and its projection onto} an OGM. Compared with the \yihui{direct} OGM baseline, which loops over all points and directly updates the occupancy state of their corresponding cells, our method achieves lower runtime, supports real-time updates and online path planning, and, like OctoMap, yields a sparse representation of large scenes.

\subsubsection{Planning Performance in Sparse Environments} In this experiment, we compared our method in terms of pathfinding success rate, path length, and \yihui{A*} and JPS planning time on two large-scale but relatively sparse point cloud scenes, \texttt{Cambridge\_15} and \texttt{Cambridge\_16}, across varying resolutions. To ensure fairness, this experiment followed the same configuration as the randomized data experiments.

\yihui{As} shown in Table \ref{tab:planning_metrics_all}, real-world results are consistent with the randomized data \yihui{results}. \yihui{POMP} shows a clear advantage in the success rate of pathfinding. Compared with the baseline direct OGM, it produces paths of comparable or shorter length, though with slightly longer planning time \yihui{in some cases}. 
\begin{figure}[h!]
    \centering
    \includegraphics[width=\linewidth]{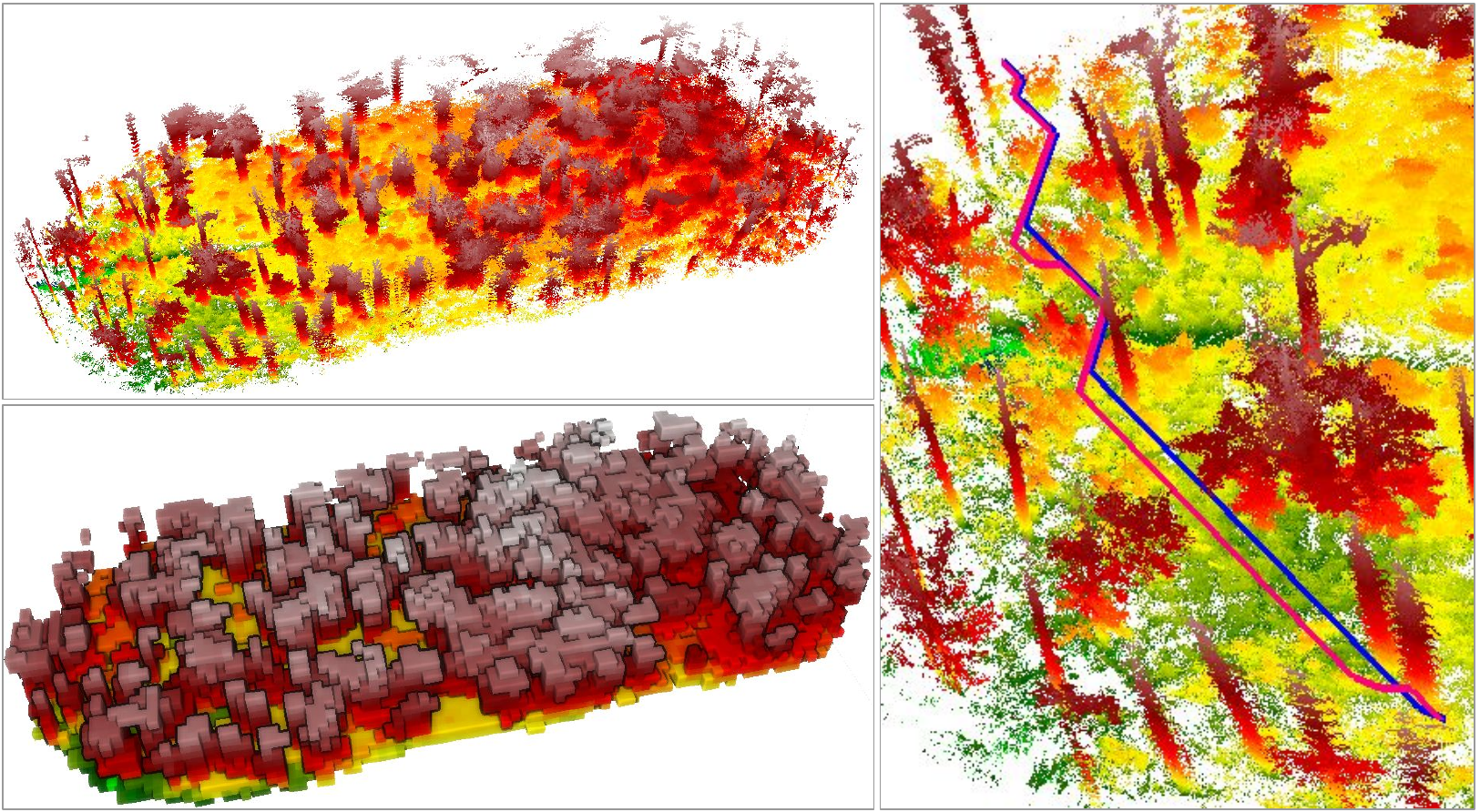}
    \caption{\texttt{Cloud\_7} from WSF-19 in the Treescope dataset. The upper-left panel shows the raw point cloud map, the lower-left panel shows the octree constructed from the point cloud, and the right panel shows the planning result. The blue path is generated by POMP, while the purple path is obtained using the direct OGM.}
    \vspace{-0.6em}
    \label{fig:Treecode}
\end{figure}

\subsubsection{Planning Performance in \yihui{Cluttered} Environments}
To study performance in \yihui{clustered} environments, we use the \texttt{Cloud\_7} (WSF-19, Treescope) \yihui{dataset}, shown in Fig. \ref{fig:Treecode}, and measure pathfinding rate, resulting path length, and A* and JPS planning time over a range of resolutions.


\yihui{As} shown in Table~\ref{tab:planning_metrics_all}, \yihui{POMP} shows a clear advantage in the success rate. Compared with the baseline direct OGM,
it produces paths of comparable or shorter length, though with slightly longer planning time. \yihui{Since POMP provides the planners with more free cells than the baseline, a slight increase in planning time is expected.}


\subsection{\yihui{Real-world experiments}}

\yihui{To evaluate the performance of POMP in a real-world setting using sensor data}, we use the processed rosbags released with Treescope. Since the Treescope dataset is composed of sensor data recorded in an uncontrolled outdoor environment, it is a better evaluation tool than in-lab hardware experimentation. The bags include LiDAR-inertial odometry and velocity-corrected point cloud sweeps produced by Faster-LIO~\cite{Bai.Xiao.Chen.ea2022}. The point clouds from the UAV Laser Scanning (ULS) sequences are captured with an Ouster OS1-64 LiDAR; specifically, we use two VAT-0723U rosbags: VAT-0723U-03, a 9.2 GiB bag spanning 380.4 seconds, and VAT-0723U-04, a 16.2 GiB bag spanning 654.4 seconds. Each LiDAR sweep contains approximately 65,536 points (64 × 1024). 


\yihui{To better emulate on-robot operation, we replay each rosbag at real-time speed with timestamps governed by the same ROS time mechanism used at runtime and run the same on-robot processing stack. Each LiDAR sweep is processed as it arrives and matched to the closest odometry and pose transform within a small time tolerance, using a fixed synchronization timeout. When the processing rate briefly falls behind the sensor rate, we prevent unbounded latency growth by using bounded queues and retaining only the most recent sweeps; any sweep that cannot be time-synchronized is dropped. The synchronized frames are then forwarded directly to the mapping backend under the same timing and latency constraints as live operation.}
\begin{figure*}[t!]
  \centering
  \includegraphics[width=\textwidth]{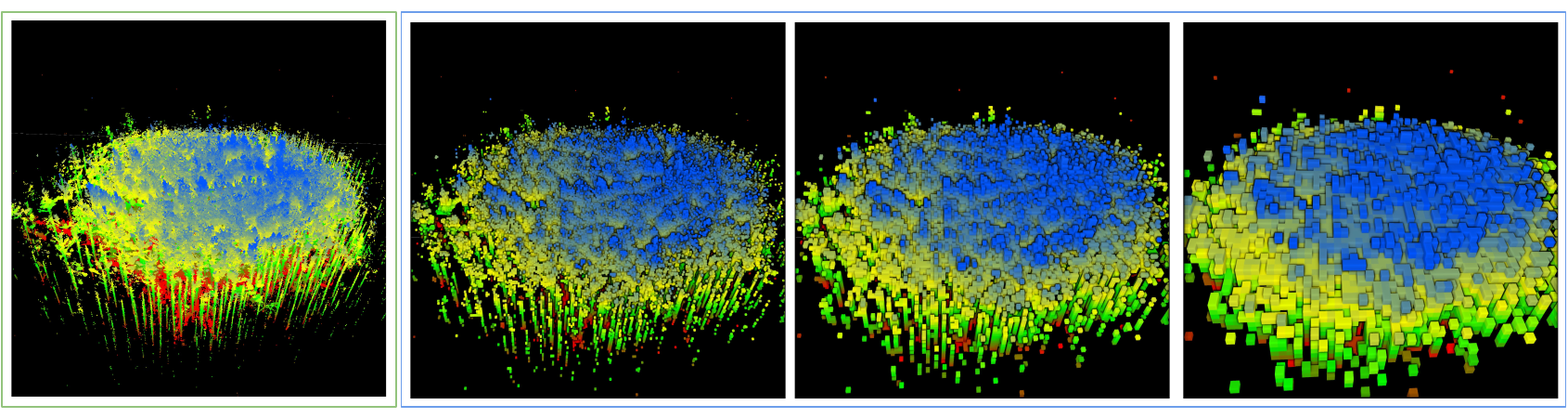}
  \caption{Visualization of the input point cloud VAT-0723U-03 and the resulting POMP Octree, from left to right with leaf sizes $r = 0.5\mathrm{m}$, $1.0\mathrm{m}$, and $2.0\mathrm{m}$.}
  \label{fig:xxx}
\end{figure*}

\begin{table}[h!]
\centering
\footnotesize
\setlength{\tabcolsep}{10.0pt}
\renewcommand{\arraystretch}{1.08}
\caption{\yihui{Experiment C-1:} Per-sweep Octree construction time (ms).}
\label{tab:mapping_build_time}

\texttt{VAT-0723U-03}
\vspace{3pt}

\begin{tabular}{l c c c}
\toprule
Res. (m) & POMP & OctoMap-MTX & OctoMap \\
\midrule
0.2 & $\textbf{1.536}\!\pm\!\textbf{0.008}$ & $5.817\!\pm\!0.096$  & $6.513\!\pm\!0.098$ \\
0.5 & $\textbf{1.333}\!\pm\!\textbf{0.009}$ & $4.592\!\pm\!0.062$ & $3.433\!\pm\!0.051$ \\
1.0 & $\textbf{1.289}\!\pm\!\textbf{0.007}$ & $4.158\!\pm\! 0.048$ & $2.601\!\pm\!0.043$ \\
1.5 & $\textbf{1.264}\!\pm\!\textbf{0.008}$ & $3.995\!\pm\!0.031$  & $2.428\!\pm\!0.030$ \\
2.0 & $\textbf{1.255}\!\pm\!\textbf{0.011}$ & $3.890\!\pm\!0.053$ & $2.291\!\pm\!0.046$ \\
2.5 & $\textbf{1.233}\!\pm\!\textbf{0.006}$ & $3.845\!\pm\!0.024$  & $2.194\!\pm\!0.050$ \\
\bottomrule
\end{tabular}

\vspace{10pt}

\texttt{VAT-0723U-04}
\vspace{3pt}

\begin{tabular}{l c c c}
\toprule
Res. (m) & POMP & OctoMap-MTX & OctoMap \\
\midrule
0.2 & $\textbf{1.521}\!\pm\!\textbf{0.021}$ & $6.233\!\pm\! 0.075$ & $6.526\!\pm\!0.078$ \\
0.5 & $\textbf{1.350}\!\pm\!\textbf{0.012}$ & $4.852\!\pm\!0.048$  & $3.559\!\pm\!0.064$ \\
1.0 & $\textbf{1.305}\!\pm\!\textbf{0.016}$ & $4.393\!\pm\!0.040$  & $2.941\!\pm\!0.063$ \\
1.5 & $\textbf{1.261}\!\pm\!\textbf{0.012}$ & $4.156\!\pm\!0.045$  & $2.657\!\pm\!0.059$ \\
2.0 & $\textbf{1.240}\!\pm\!\textbf{0.011}$ & $3.997\!\pm\! 0.056$  & $2.643\!\pm\!0.040$ \\
2.5 & $\textbf{1.239}\!\pm\!\textbf{0.008}$ & $3.941\!\pm\!0.028$  & $2.597\!\pm\!0.056$ \\
\bottomrule
\end{tabular}

\end{table}

\subsubsection{Octree Construction}

The per-sweep Octree construction time of POMP is evaluated against that of OctoMap-MTX and the original OctoMap across different leaf-node sizes (i.e., map resolutions). For original OctoMap, we used the official open-source implementation available on GitHub~\cite{hornung2013octomap}. To align with the default setting in the OctoMap source code, we use the same 16-level tree with leaf resolution $r$. This configuration yields a maximum spatial extent of $r \cdot 2^{15}$. Although our method can adapt the tree depth based on the estimated scene size, we keep the depth fixed here for a fair comparison.




As shown in Table~\ref{tab:mapping_build_time}, we replay each rosbag at its nominal rate and measure per-sweep construction time over the full playback. Each configuration is repeated 10 times, and we report the mean and standard deviation. Overall, \yihui{POMP} achieves lower runtime with reduced variability, and is typically 2-3xfaster than OctoMap and OctoMap-MTX, meeting the real-time requirements of a lightweight online mapping pipeline.

\subsubsection{Occupancy Grid Construction from Octree}
\yihui{The time taken to construct occupancy grids from the POMP octree and the original OctoMap octree is compared.} For each setting, we run the conversion 500 times on the constructed Octree. As the mean and standard deviation in Table~\ref{tab:mapping_build_time_octomap} indicate, POMP achieves lower average \yihui{conversion} time and lower variability.

\begin{table}[h!]
\centering
\footnotesize
\setlength{\tabcolsep}{2.pt}
\renewcommand{\arraystretch}{1.05}
\caption{\yihui{Experiment C-2:} Per-sweep mapping time (ms).}
\label{tab:mapping_build_time_octomap}

\begin{tabular}{l rr rr}
\toprule
& \multicolumn{2}{c}{\texttt{VAT-0723U-03}}
& \multicolumn{2}{c}{\texttt{VAT-0723U-04}} \\
\cmidrule(lr){2-3}\cmidrule(lr){4-5}
Res. (m)
& \multicolumn{1}{c}{POMP } & \multicolumn{1}{c}{OctoMap}
& \multicolumn{1}{c}{POMP} & \multicolumn{1}{c}{OctoMap} \\
\midrule
0.2
& $\textbf{11.803}\!\pm\!\textbf{0.448}$ & $52.364\!\pm\!1.643$
& $\textbf{13.245}\!\pm\!\textbf{0.426}$ & $59.363\!\pm\!1.530$ \\
0.5
& $\textbf{2.735}\!\pm\!\textbf{0.229}$  & $9.136\!\pm\!0.366$
& $\textbf{2.929}\!\pm\!\textbf{0.284}$  & $9.569\!\pm\!0.482$ \\
1.0
& $\textbf{0.429}\!\pm\!\textbf{0.071}$  & $1.573\!\pm\!0.157$
& $\textbf{0.465}\!\pm\!0.288$            & $1.650\!\pm\!\textbf{0.191}$ \\
1.5
& $\textbf{0.192}\!\pm\!\textbf{0.069}$  & $0.468\!\pm\!0.082$
& $\textbf{0.189}\!\pm\!\textbf{0.076}$  & $0.475\!\pm\!0.079$ \\
2.0
& $\textbf{0.127}\!\pm\!0.066$            & $0.238\!\pm\!\textbf{0.059}$
& $\textbf{0.136}\!\pm\!0.062$            & $0.236\!\pm\!\textbf{0.058}$ \\
2.5
& $\textbf{0.098}\!\pm\!0.050$            & $0.140\!\pm\!\textbf{0.037}$
& $\textbf{0.102}\!\pm\!0.058$            & $0.139\!\pm\!\textbf{0.040}$ \\
\bottomrule
\end{tabular}

\vspace{4pt}
\end{table}

\begin{table*}[t]
\centering
\footnotesize
\setlength{\tabcolsep}{2.0pt}
\renewcommand{\arraystretch}{1.12}
\caption{\yihui{Experiment C-3:} Planning metrics across resolutions.}
\label{tab:planning_metrics_side_by_side}

\begin{tabular}{l rr !{\hskip 1.8pt\vrule width 0.1pt\hskip 1.6pt} rr !{\hskip 1.8pt\vrule width 0.1pt\hskip 1.6pt} rr !{\hskip 1.8pt\vrule width 0.1pt\hskip 1.6pt} rr !{\hskip 1.8pt\vrule width 1.0pt\hskip 1.6pt} rr !{\hskip 1.8pt\vrule width 0.1pt\hskip 1.6pt} rr !{\hskip 1.8pt\vrule width 0.1pt\hskip 1.6pt} rr!{\hskip 1.8pt\vrule width 0.1pt\hskip 1.6pt} rr}
\toprule
& \multicolumn{8}{c}{\texttt{VAT-0723U-03}} & \multicolumn{8}{c}{\texttt{VAT-0723U-04}} \\
\cmidrule(lr){2-9}\cmidrule(lr){10-17}
Res. (m)
& \multicolumn{2}{c}{Succ. (\%) $\uparrow$}
& \multicolumn{2}{c}{A* time (ms) $\downarrow$}
& \multicolumn{2}{c}{JPS time (ms) $\downarrow$}
& \multicolumn{2}{c}{Length (m) $\downarrow$}
& \multicolumn{2}{c}{Succ. (\%) $\uparrow$}
& \multicolumn{2}{c}{A* time (ms) $\downarrow$}
& \multicolumn{2}{c}{JPS time (ms) $\downarrow$}
& \multicolumn{2}{c}{Length (m) $\downarrow$} \\
\cmidrule(lr){2-3}\cmidrule(lr){4-5}\cmidrule(lr){6-7}\cmidrule(lr){8-9}
\cmidrule(lr){10-11}\cmidrule(lr){12-13}\cmidrule(lr){14-15}\cmidrule(lr){16-17}
& \multicolumn{1}{c}{POMP} & \multicolumn{1}{c}{OGM}
& \multicolumn{1}{c}{POMP} & \multicolumn{1}{c}{OGM}
& \multicolumn{1}{c}{POMP} & \multicolumn{1}{c}{OGM}
& \multicolumn{1}{c}{POMP} & \multicolumn{1}{c}{OGM}
& \multicolumn{1}{c}{POMP} & \multicolumn{1}{c}{OGM}
& \multicolumn{1}{c}{POMP} & \multicolumn{1}{c}{OGM}
& \multicolumn{1}{c}{POMP} & \multicolumn{1}{c}{OGM}
& \multicolumn{1}{c}{POMP} & \multicolumn{1}{c}{OGM} \\
\midrule
0.2
& $\textbf{90.60}\%$ & $88.80\%$
& $52.833$ & $\textbf{52.200}$
& $43.766$ & $\textbf{42.787}$
& $\textbf{15.466}$ & $15.469$
& $\textbf{88.20}\%$ & $86.00\%$
& $51.005$ & $\textbf{50.810}$
& $41.391$ & $41.391$
& $\textbf{15.015}$ & $15.019$ \\
0.5
& $\textbf{69.00}\%$ & $65.00\%$
& $\textbf{1.100}$ & $1.405$
& $\textbf{0.571}$ & $0.585$
& $\textbf{15.491}$ & $15.507$
& $\textbf{73.60}\%$ & $69.60\%$
& $\textbf{1.053}$ & $1.355$
& $\textbf{0.579}$ & $0.591$
& $\textbf{14.594}$ & $14.637$ \\
1.0
& $\textbf{44.00}\%$ & $38.40\%$
& $\textbf{0.113}$ & $0.153$
& $0.069$ & $\textbf{0.068}$
& $\textbf{14.303}$ & $14.389$
& $\textbf{42.80}\%$ & $37.00\%$
& $\textbf{0.124}$ & $0.161$
& $\textbf{0.074}$ & $0.076$
& $\textbf{15.207}$ & $15.262$ \\
1.5
& $\textbf{27.80}\%$ & $24.20\%$
& $\textbf{0.050}$ & $0.067$
& $\textbf{0.033}$ & $0.034$
& $\textbf{16.427}$ & $16.740$
& $\textbf{26.60}\%$ & $20.20\%$
& $\textbf{0.048}$ & $0.053$
& $\textbf{0.032}$ & $0.033$
& $\textbf{16.024}$ & $16.247$ \\
2.0
& $\textbf{12.80}\%$ & $8.60\%$
& $0.041$ & $\textbf{0.029}$
& $0.018$ & $0.018$
& $\textbf{16.950}$ & $18.161$
& $\textbf{11.40}\%$ & $7.60\%$
& $0.037$ & $\textbf{0.024}$
& $0.015$ & $0.015$
& $\textbf{13.689}$ & $14.182$ \\
2.5
& $\textbf{3.40}\%$ & $1.40\%$
& $\textbf{0.016}$ & $0.019$
& $0.012$ & $\textbf{0.009}$
& $\textbf{21.576}$ & $23.684$
& $\textbf{2.60}\%$ & $1.80\%$
& $\textbf{0.011}$ & $0.014$
& $0.009$ & $\textbf{0.007}$
& $\textbf{13.604}$ & $13.780$ \\
\bottomrule
\end{tabular}

\vspace{4pt}
\parbox{\textwidth}{\scriptsize
\emph{Note:} Bold indicates the better value between POMP and OGM for each bag and resolution. Success rate denotes the pathfinding rate, and length denotes the path length.
}
\end{table*}

\subsubsection{\yihui{Navigability and Planning}}
We also \yihui{repeat Experiments A-3 and A-4  for these data.} Specifically, we compare (i) the navigability of the occupancy grids converted from the \yihui{POMP Octree} (Table~\ref{tab:nsr_mean}) and (ii) the planning performance on these occupancy grids (Table~\ref{tab:planning_metrics_side_by_side}) against occupancy grids obtained from original OctoMap. Planning performance is evaluated using both A* and JPS, including pathfinding success rate, path length, and planning time. Since JPS is a pruned, accelerated variant of A*, it returns the same shortest-path length as A* on the same grid (when using the same connectivity and edge costs). Therefore, A* and JPS have identical path lengths in our experiments, and we do not report them separately.

\begin{table}[h!]
\centering
\footnotesize
\setlength{\tabcolsep}{8.pt}
\renewcommand{\arraystretch}{1.05}
\caption{\yihui{Experiment C-4:} Per-sweep Navigable Space Ratio (\%).}
\label{tab:nsr_mean}

\begin{tabular}{l rr rr}
\toprule
& \multicolumn{2}{c}{\texttt{VAT-0723U-03}}
& \multicolumn{2}{c}{\texttt{VAT-0723U-04}} \\
\cmidrule(lr){2-3}\cmidrule(lr){4-5}
Res. (m)
& \multicolumn{1}{c}{POMP} & \multicolumn{1}{c}{OGM}
& \multicolumn{1}{c}{POMP} & \multicolumn{1}{c}{OGM} \\
\midrule
0.2  & $\textbf{94.17}\%$ & $93.44\%$
     & $\textbf{94.11}\%$ & $93.41\%$ \\
0.5  & $\textbf{84.20}\%$ & $82.08\%$
     & $\textbf{84.02}\%$ & $81.92\%$ \\
1.0  & $\textbf{66.38}\%$ & $62.35\%$
     & $\textbf{65.23}\%$ & $61.41\%$ \\
1.5  & $\textbf{50.42}\%$ & $45.18\%$
     & $\textbf{49.13}\%$ & $43.40\%$ \\
2.0  & $\textbf{35.01}\%$ & $27.71\%$
     & $\textbf{32.94}\%$ & $26.53\%$ \\
2.5  & $\textbf{23.50}\%$ & $15.25\%$
     & $\textbf{24.00}\%$ & $18.25\%$ \\
\bottomrule
\end{tabular}

\end{table}

\yihui{As shown in} Table~\ref{tab:nsr_mean}, our maps preserve more navigable free space than the \yihui{direct OGM} method across resolutions; consequently, as reported in Table~\ref{tab:planning_metrics_side_by_side}, \yihui{POMP} achieves a higher pathfinding success rate. The improvement holds at fine resolutions (where finding a path is usually easier) and at coarse resolutions (where planning often fails). At the resolutions that best match the scene and the planner, the improvement becomes greater, which is consistent with what our method is designed to do.

\section{DISCUSSION}
\yihui{In our mapping to planning pipeline, we re-partition a fixed-resolution occupancy grid map using OctoMap leaf boundaries at depth $n$, then apply occupancy thresholding with a conservative safety margin and assign each cell's navigability via Diagonal Examination (Fig.~\ref{fig:ogm_nav}). The key intuition is that uniform partitioning improves coverage, making gaps near the grid resolution less likely to be missed. Given the scene scale, we can further choose an appropriate OctoMap tree depth to ensure sufficient spatial coverage while controlling memory and computation.}

In implementation, map building is integrated into the planning pipeline, allowing mapping to be performed during planning. \yihui{The Octree insertion and update operations, including the safety-state updates, are parallelized. The subsequent octree to OGM projection is also parallelized. With dense point clouds, POMP is typically faster than a naive point-wise serial loop that writes directly into an OGM, meeting the requirements of synchronous online operation. 
Compared with the serial Octree-to-OGM conversion in the original method, POMP’s octree to OGM conversion incurs little additional computational cost; the proposed region safety state update does not introduce excessive cost. In the serial baseline, each leaf node typically maps to a single OGM cell update; in POMP, each leaf node is split into four regions in 2D and eight in 3D, and navigability is written to the OGM via one-way diagonal pair updates (a diagonal pair of cells is updated together), requiring two pairs in 2D and four pairs in 3D. This procedure can be further accelerated via parallel execution, while atomic operations prevent race conditions under concurrent updates. Moreover, given a desired OGM configuration (resolution, origin, and spatial bounds) based on the expected scene size, the corresponding octree placement (origin and extent) can be determined so that the octree is instantiated with the appropriate size, location, and depth to match the target OGM layout.}

We validate POMP through extensive benchmarks on diverse real-world datasets, including LiDAR and reconstruction maps spanning outdoor urban scenes, indoor rooms, and forested environments. Additional tests use randomized scene configurations with diverse geometric combinations. Experiments cover point clouds of varying scales and densities, from sparse to dense, and include realistic rosbag replays that emulate online operation.

Evaluation considers a broad set of metrics. Across the above scenes, multiple point cloud sizes, and multiple resolutions, we measure OctoMap build time, octree to OGM projection time, and navigability differences relative to a standard OGM baseline. Planning performance is also reported across resolutions and scenarios, including pathfinding success rate, search time, and path length under multiple search-based planning methods, with an additional experiment evaluating the impact of the threshold on success rate.

To verify consistency with the standard serial implementation, we examine the proposed POMP construction procedure from both analytical and empirical perspectives. In particular, we compare the resulting data structures and validate point cloud assignments at the node level to confirm equivalence. To demonstrate construction efficiency, we further benchmark against state-of-the-art tree based structure building methods, where our approach consistently achieves faster build time.

\section{CONCLUSIONS}

This article presents a framework for fixed-resolution free-
space refinement and efficient octree-based occupancy-grid
mapping. By accelerating dense point-cloud insertion and
occupancy-grid-map updates without altering the underlying
data structure or introducing additional computational over-
head, the framework provides an efficient basis for
search-based planning. The framework builds on three core techniques.


First, octree construction is accelerated through parallel multithreaded computation. Because concurrent updates in a hierarchical tree are nontrivial, compare-and-swap with atomic pointers ensures safe node creation and updates, while atomic operations propagate leaf states and point-cloud updates to a planner-facing occupancy grid without race conditions. This combination achieves mapping time comparable to a point-wise OGM insertion baseline and outperforms that baseline as point clouds become denser.

Second, discretization accuracy is improved without further subdividing the tree. Region-level safety states are stored in a compact byte-level representation, avoiding the memory overhead of deeper trees while inducing a cell-level safety margin. \yihui{This region-level encoding} refines free-space representation at a fixed resolution and improves search-based planning efficiency.

Third, a hybrid map couples mapping and planning. An \yihui{Octree}-backed representation is maintained alongside a planner-facing occupancy grid, enabling synchronous operation and achieving higher pathfinding success rates and comparable or shorter paths than conventional fixed-resolution OGM pipelines.

In summary, POMP shows that significant gains in planning performance can be achieved not by globally refining map resolution, but by more effectively exploiting free space within each fixed-resolution cell. Through parallel octree construction, compact region-level encoding, and a planner-facing hybrid map, POMP brings coarse-resolution planning closer to fine-resolution performance while retaining the efficiency advantages of coarse maps. As a result, it provides a practical path toward faster, more memory-efficient, and more reliable planning in dense and cluttered environments, and can be readily integrated into existing search-based autonomy pipelines.
\label{app:additional_experimental_details}

\small
\bibliographystyle{IEEEtran}
\bibliography{IEEEtranBST/IEEEexample}

@article{hornung2013octomap,
  author={Hornung, Armin and Wurm, Kai M and Bennewitz, Maren and Stachniss, Cyrill and Burgard, Wolfram},
  title={OctoMap: An efficient probabilistic {3D} mapping framework based on octrees},
  journal={Auton. Robots},
  year={2013},
  volume={34},
  pages={189--206},
  publisher={Springer},
}

@article{liu2023dataframe,
  author    = {Liu, Zhenyu and {van Oosterom}, Peter and Balado, Jes{\'u}s and Swart, Arjen and Beers, Bart},
  title     = {Data frame aware optimized Octomap-based dynamic object detection and removal in Mobile Laser Scanning data},
  journal   = {Alexandria Engineering Journal},
  year      = {2023},
  volume    = {74},
  pages     = {327--344},
  publisher = {Elsevier},
}

@article{sun2018recurrentoctomap,
  author={Sun, Li and Yan, Zhi and Zaganidis, Anestis and Zhao, Cheng and Duckett, Tom},
  title={{Recurrent-OctoMap}: Learning State-Based Map Refinement for Long-Term Semantic Mapping With 3-D LiDAR Data},
  journal={IEEE Robot. Autom. Lett.},
  year={2018},
  volume={3},
  pages={3749--3756},
  publisher={IEEE},
}

@InProceedings{Chen.Li.Wan.ea2025,
  author    = {Chen, Peiqing and Li, Minghao and Wan, Zishen and Hsiao, Yu-Shun and Yu, Minlan and Reddi, Vijay Janapa and Liu, Zaoxing},
  booktitle = {Proc. ACM Int. Conf. Archit. Support Program. Lang. Oper. Syst.},
  title     = {{OctoCache}: Caching Voxels for Accelerating {3D} Occupancy Mapping in Autonomous Systems},
  year      = {2025},
  pages     = {704--718},
  doi       = {10.1145/3676641.3716263},
  publisher = {ACM},
  url       = {https://doi.org/10.1145/3676641.3716263},
}

@InProceedings{curless1996volumetric,
  author    = {Curless, Brian and Levoy, Marc},
  title     = {A volumetric method for building complex models from range images},
  booktitle = {Proc. ACM SIGGRAPH},
  year      = {1996},
  pages     = {303--312},
}

@InProceedings{SCC.DeGregorio.DiStefano2017,
  author       = {De Gregorio, Daniele and Di Stefano, Luigi},
  booktitle    = {Proc. IEEE Int. Conf. Robot. Autom.},
  title        = {{SkiMap}: An efficient mapping framework for robot navigation},
  year         = {2017},
  organization = {IEEE},
  pages        = {2569--2576},
  doi          = {10.1109/ICRA.2017.7989299},
}

@InProceedings{oleynikova2017voxblox,
  author    = {Oleynikova, Helen and Taylor, Zachary and Fehr, Marius and Siegwart, Roland and Nieto, Juan},
  title     = {Voxblox: Incremental {3D} Euclidean Signed Distance Fields for On-Board MAV Planning},
  booktitle = {Proc. IEEE/RSJ Int. Conf. Intell. Robots Syst. (IROS)},
  year      = {2017},
  pages     = {1366--1373},
}

@article{duberg2020ufomap,
  author={Duberg, Daniel and Jensfelt, Patric},
  title={{UFOMap}: An efficient probabilistic {3D} mapping framework that embraces the unknown},
  journal={IEEE Robot. Autom. Lett.},
  year={2020},
  volume={5},
  number={4},
  pages={6411--6418},
  publisher={IEEE},
}

@article{cai2024occupancy,
  author={Cai, Yixi and Kong, Fanze and Ren, Yunfan and Zhu, Fangcheng and Lin, Jiarong and Zhang, Fu},
  title={Occupancy Grid Mapping Without Ray-Casting for High-Resolution {LiDAR} Sensors},
  journal={IEEE Trans. Robot.},
  year={2024},
  volume={40},
  pages={172--192},
  publisher={IEEE},
}

@InProceedings{Pan.Kompis.Bartolomei.ea2022,
  author    = {Pan, Yue and Kompis, Yves and Bartolomei, Luca and Mascaro, Ruben and Stachniss, Cyrill and Chli, Margarita},
  booktitle = {Proc. IEEE/RSJ Int. Conf. Intell. Robots Syst.},
  title     = {{Voxfield}: Non-Projective Signed Distance Fields for Online Planning and {3D} Reconstruction},
  year      = {2022},
  pages     = {5331--5338},
  doi       = {10.1109/IROS47612.2022.9981318},
  publisher = {IEEE},
  url       = {https://ieeexplore.ieee.org/document/9981318},
}

@Article{7839930,
  author  = {Liu, Sikang and Watterson, Michael and Mohta, Kartik and Sun, Ke and Bhattacharya, Subhrajit and Taylor, Camillo J. and Kumar, Vijay},
  title   = {Planning Dynamically Feasible Trajectories for Quadrotors Using Safe Flight Corridors in 3-{D} Complex Environments},
  journal = {IEEE Robot. Autom. Lett.},
  year    = {2017},
  volume  = {2},
  number  = {3},
  pages   = {1688--1695},
  doi     = {10.1109/LRA.2017.2663526},
}

@InProceedings{Han.Gao.Zhou.ea2019,
  author    = {Han, Luxin and Gao, Fei and Zhou, Boyu and Shen, Shaojie},
  booktitle = {Proc. IEEE/RSJ Int. Conf. Intell. Robots Syst.},
  title     = {{FIESTA}: Fast Incremental Euclidean Distance Fields for Online Motion Planning of Aerial Robots},
  year      = {2019},
  pages     = {4423--4430},
  doi       = {10.1109/IROS40897.2019.8968199},
  url       = {https://ieeexplore.ieee.org/document/8968199},
}

@InProceedings{SCC.Qureshi.Ogri.ea2024,
  author       = {Qureshi, Muzaffar and Ogri, Tochukwu Elijah and Bell, Zachary I. and Kamalapurkar, Rushikesh},
  booktitle    = {Proc. IEEE Conf. Control Technol. Appl.},
  title        = {Scalar field mapping with adaptive high-intensity region avoidance},
  year         = {2024},
  address      = {Newcastle upon Tyne, UK},
  month        = aug,
  pages        = {388--393},
  doi          = {10.1109/CCTA60707.2024.10666509},
  projects     = {afrl-fa8651-23-1-0006},
  url          = {https://ieeexplore.ieee.org/document/10666509},
  url_preprint = {https://scc-lab.github.io/Preprints/},
}

@Article{zhou2020ego,
  author    = {Zhou, Xin and Wang, Zhepei and Ye, Hongkai and Xu, Chao and Gao, Fei},
  title     = {Ego-planner: An {ESDF}-free gradient-based local planner for quadrotors},
  journal   = {IEEE Robot. Autom. Lett.},
  year      = {2020},
  volume    = {6},
  number    = {2},
  pages     = {478--485},
  publisher = {IEEE},
}

@InProceedings{liu.mao.belta2025acc,
  author    = {Liu, S. and Mao, Y. and Belta, C. A.},
  title     = {Safety-Critical Planning and Control for Dynamic Obstacle Avoidance Using Control Barrier Functions},
  booktitle = {Proc. Am. Control Conf. (ACC)},
  year      = {2025},
  pages     = {348--354},
  address   = {Denver, CO, USA},
  doi       = {10.23919/ACC63710.2025.11107805},
}

@Article{niessner2013real,
  author    = {Nie{\ss}ner, Matthias and Zollh{\"o}fer, Michael and Izadi, Shahram and Stamminger, Marc},
  title     = {Real-time {3D} reconstruction at scale using voxel hashing},
  journal   = {ACM Transactions on Graphics},
  year      = {2013},
  volume    = {32},
  number    = {6},
  pages     = {1--11},
  publisher = {ACM New York, NY, USA},
}

@Book{reinders2007intel,
  title     = {Intel threading building blocks: outfitting C++ for multi-core processor parallelism},
  publisher = {O'Reilly Media, Inc.},
  year      = {2007},
  author    = {Reinders, James},
}

@InProceedings{6599048,
  author    = {Keller, Maik and Lefloch, Damien and Lambers, Martin and Izadi, Shahram and Weyrich, Tim and Kolb, Andreas},
  title     = {Real-Time {3D} Reconstruction in Dynamic Scenes Using Point-Based Fusion},
  booktitle = {Proc. Int. Conf. 3D Vision (3DV)},
  year      = {2013},
  pages     = {1--8},
  doi       = {10.1109/3DV.2013.9},
}

@InProceedings{KinectFusion,
  author    = {Newcombe, Richard A. and Izadi, Shahram and Hilliges, Otmar and Molyneaux, David and Kim, David and Davison, Andrew J. and Kohi, Pushmeet and Shotton, Jamie and Hodges, Steve and Fitzgibbon, Andrew},
  title     = {KinectFusion: Real-time dense surface mapping and tracking},
  booktitle = {Proc. IEEE Int. Symp. Mixed Augmented Reality},
  year      = {2011},
  pages     = {127--136},
  doi       = {10.1109/ISMAR.2011.6092378},
}

@Article{thrun2002probabilistic,
  author  = {Thrun, Sebastian},
  title   = {Probabilistic robotics},
  journal = {Communications of the ACM},
  year    = {2002},
  volume  = {45},
  number  = {3},
  pages   = {52--57},
  publisher = {ACM New York, NY, USA},
}

@incollection{dijkstra2022note,
  title={A note on two problems in connexion with graphs},
  author={Dijkstra, Edsger W},
  booktitle={Edsger Wybe Dijkstra: his life, work, and legacy},
  pages={287--290},
  year={2022}
}

@Article{hart1968formal,
  author    = {Hart, Peter E. and Nilsson, Nils J. and Raphael, Bertram},
  journal   = {IEEE Trans. Syst. Sci. Cybern.},
  title     = {A formal basis for the heuristic determination of minimum cost paths},
  year      = {1968},
  volume    = {4},
  number    = {2},
  pages     = {100--107},
  publisher = {IEEE},
}

@InProceedings{SCC.Zhu.Li.ea2024,
  author    = {Jun Zhu and Hongyi Li and Zhepeng Wang and Shengjie Wang and Tao Zhang},
  booktitle = {Proc. IEEE Int. Conf. Robot. Autom.},
  title     = {{i-Octree}: A Fast, Lightweight, and Dynamic Octree for Proximity Search},
  year      = {2024},
  pages     = {12290--12296},
  doi       = {10.1109/ICRA57147.2024.10611019},
}

@InProceedings{SCC.Min.Han.ea2021accel,
  author       = {Min, Heajung and Han, Kyung Min and Kim, Young J.},
  booktitle    = {Proc. IEEE Int. Conf. Robot. Autom.},
  title        = {Accelerating Probabilistic Volumetric Mapping using Ray-Tracing Graphics Hardware},
  year         = {2021},
  organization = {IEEE},
  pages        = {5440--5445},
}

@InProceedings{SCC.Rusu.Cousins.ea2011,
  author    = {Radu Bogdan Rusu and Steve Cousins},
  booktitle = {Proc. IEEE Int. Conf. Robot. Autom.},
  title     = {{3D} is here: Point Cloud Library ({PCL})},
  year      = {2011},
  pages     = {1--4},
  doi       = {10.1109/ICRA.2011.5980567},
}

@InProceedings{stentz1994optimal,
  author    = {Stentz, Anthony},
  title     = {Optimal and efficient path planning for partially-known environments},
  booktitle = {Proc. IEEE Int. Conf. Robot. Autom.},
  year      = {1994},
  pages     = {3310--3317},
}

@InProceedings{harabor2011online,
  author    = {Harabor, Daniel and Grastien, Alban},
  title     = {Online graph pruning for pathfinding on grid maps},
  booktitle = {Proc. AAAI Conf. Artif. Intell.},
  year      = {2011},
  volume    = {25},
  number    = {1},
  pages     = {1114--1119},
}

@Article{replica19arxiv,
  author  = {Straub, Julian and Whelan, Thomas and Ma, Lingni and Chen, Yufan and Wijmans, Erik and Green, Simon and Engel, Jakob J. and Mur-Artal, Raul and Ren, Carl and Verma, Shobhit and Clarkson, Anton and Yan, Mingfei and Budge, Brian and Yan, Yajie and Pan, Xiaqing and Yon, June and Zou, Yuyang and Leon, Kimberly and Carter, Nigel and Briales, Jesus and Gillingham, Tyler and Mueggler, Elias and Pesqueira, Luis and Savva, Manolis and Batra, Dhruv and Strasdat, Hauke M. and De Nardi, Renzo and Goesele, Michael and Lovegrove, Steven and Newcombe, Richard},
  title   = {The {R}eplica Dataset: A Digital Replica of Indoor Spaces},
  journal = {arXiv preprint arXiv:1906.05797},
  year    = {2019},
}

@Article{hu2022sensaturban,
  author    = {Hu, Qingyong and Yang, Bo and Khalid, Sheikh and Xiao, Wen and Trigoni, Niki and Markham, Andrew},
  title     = {Sensat{U}rban: Learning semantics from urban-scale photogrammetric point clouds},
  journal   = {Int. J. Comput. Vis.},
  year      = {2022},
  volume    = {130},
  number    = {2},
  pages     = {316--343},
  publisher = {Springer},
}

@InProceedings{SCC.Durvasula.Kiguru.ea2023,
  author       = {Durvasula, Sankeerth and Kiguru, Raymond and Mathur, Samarth and Xu, Jenny and Lin, Jimmy and Vijaykumar, Nandita},
  booktitle    = {Proc. Int. Conf. Parallel Archit. Compilation Tech.},
  title        = {{VoxelCache}: Accelerating Online Mapping in Robotics and 3{D} Reconstruction Tasks},
  year         = {2023},
  organization = {ACM},
  pages        = {239--251},
  doi          = {10.1145/3559009.3569675},
}

@Article{SCC.Kwon.Kim.ea2019,
  author  = {Kwon, Youngsun and Kim, Donghyuk and An, Inkyu and Yoon, Sung-eui},
  title   = {Super Rays and Culling Region for Real-Time Updates on Grid-Based Occupancy Maps},
  journal = {IEEE Trans. Robot.},
  year    = {2019},
  volume  = {35},
  number  = {2},
  pages   = {482--497},
  doi     = {10.1109/TRO.2018.2889262},
}

@Article{Bai.Xiao.Chen.ea2022,
  author  = {Bai, Chunge and Xiao, Tao and Chen, Yajie and Wang, Haoqian and Zhang, Fang and Gao, Xiang},
  title   = {{Faster-LIO}: Lightweight Tightly Coupled Lidar-Inertial Odometry Using Parallel Sparse Incremental Voxels},
  journal = {IEEE Robot. Autom. Lett.},
  year    = {2022},
  volume  = {7},
  number  = {2},
  pages   = {4861--4868},
  doi     = {10.1109/LRA.2022.3152830},
}

@book{latombe1991robot,
  title     = {Robot Motion Planning},
  publisher = {Kluwer Academic Publishers},
  author    = {Latombe, Jean-Claude},
  year      = {1991},
  address   = {Boston, MA}
}

@Misc{SCC.Jia.Yang.ea2022,
  author           = {Jia, Tianyu and Yang, En-Yu and Hsiao, Yu-Shun and Cruz, Jonathan and Brooks, David and Wei, Gu-Yeon and Reddi, Vijay Janapa},
  howpublished     = {{arXiv:2205.03325}},
  title            = {{OMU}: A Probabilistic 3{D} Occupancy Mapping Accelerator for Real-time {O}cto{M}ap at the Edge},
  year             = {2022},
}

@Misc{cheng2023treescope,
  author       = {Derek Cheng and Fernando Cladera Ojeda and Ankit Prabhu and Xu Liu and Alan Zhu and Patrick Corey Green and Reza Ehsani and Pratik Chaudhari and Vijay Kumar},
  title        = {TreeScope: An Agricultural Robotics Dataset for LiDAR-Based Mapping of Trees in Forests and Orchards},
  howpublished = {{arXiv}:2310.02162},
  year         = {2023},
}

@Article{30720,
  author  = {Elfes, A.},
  title   = {Using occupancy grids for mobile robot perception and navigation},
  journal = {Computer},
  year    = {1989},
  volume  = {22},
  number  = {6},
  pages   = {46--57},
  doi     = {10.1109/2.30720},
}

@InProceedings{1087316,
  author    = {Moravec, H. and Elfes, A.},
  title     = {High resolution maps from wide angle sonar},
  booktitle = {Proc. IEEE Int. Conf. Robot. Autom.},
  year      = {1985},
  volume    = {2},
  pages     = {116--121},
  doi       = {10.1109/ROBOT.1985.1087316},
}

@Misc{SCC.Cai.Xu.ea2021,
  author       = {Yixi Cai and Wei Xu and Fu Zhang},
  title        = {{ikd-Tree}: An Incremental {K-D} Tree for Robotic Applications},
  howpublished = {{arXiv}:2102.10808},
  year         = {2021},
}

@InProceedings{SCC.Florence.Carter.ea2018,
  author       = {Florence, Peter R. and Carter, John and Ware, Jake and Tedrake, Russ},
  booktitle    = {Proc. IEEE Int. Conf. Robot. Autom.},
  title        = {{Nano{M}ap}: Fast, Uncertainty-Aware Proximity Queries with Lazy Search Over Local 3{D} Data},
  year         = {2018},
  organization = {IEEE},
  pages        = {7631--7638},
}

@Article{SCC.Min.Han.ea2023.OctoMap-RT,
  author  = {Min, Heajung and Han, Kyung Min and Kim, Young J.},
  title   = {{OctoMap-RT}: Fast Probabilistic Volumetric Mapping Using Ray-Tracing {GPU}s},
  journal = {IEEE Robot. Autom. Lett.},
  year    = {2023},
  volume  = {8},
  number  = {9},
  pages   = {5696--5703},
  doi     = {10.1109/LRA.2023.3300227},
}

\end{document}